%% file: arxiv.tex
\documentclass{article} % For LaTeX2e
\usepackage{arxiv,times}

\input{math_commands.tex}

\usepackage{hyperref}
\usepackage{url}
\usepackage{graphicx}
\usepackage{amsmath}
\usepackage{booktabs}
\usepackage{multirow}
\usepackage{adjustbox}

\title{UniMMVSR: A Unified Multi-Modal Framework for Cascaded Video Super-Resolution}

\author{\hspace{30pt} Shian Du\textsuperscript{$\rm 1^{\ast}$}, Menghan Xia\textsuperscript{$\rm 2^{\dagger}$}, Chang Liu\textsuperscript{\rm 1}, Quande Liu\textsuperscript{\rm 3}, Xintao Wang\textsuperscript{\rm 3}, \\
\hspace{120pt} \textbf{Pengfei Wan}\textsuperscript{\rm 3}, \textbf{Xiangyang Ji}\textsuperscript{$\rm 1^{\dagger}$} \\ 
\hspace{45pt} \textsuperscript{\rm 1}Tsinghua University, \textsuperscript{\rm 2}Huazhong University of Science and Technology, \\ 
\hspace{105pt} \textsuperscript{\rm 3}Kling Team, Kuaishou Technology \\
\hspace{75pt} \textcolor{cyan}{\url{https://shiandu.github.io/UniMMVSR-website/}} \\
}

\newcommand\nnfootnote[1]{%
  \begin{NoHyper}
  \renewcommand\thefootnote{}\footnote{#1}%
  \addtocounter{footnote}{-1}%
  \end{NoHyper}
}

\usepackage{ulem}

\iclrfinalcopy % Uncomment for camera-ready version, but NOT for submission.
\begin{document}

\maketitle
\nnfootnote{$\ast$ This work was conducted during the author's internship at Kling Team, Kuaishou Technology.}
\nnfootnote{$\dagger$ Corresponding author.}

\begin{figure}[h]
\vspace{-4.0em}
\includegraphics[width=1.0\linewidth]{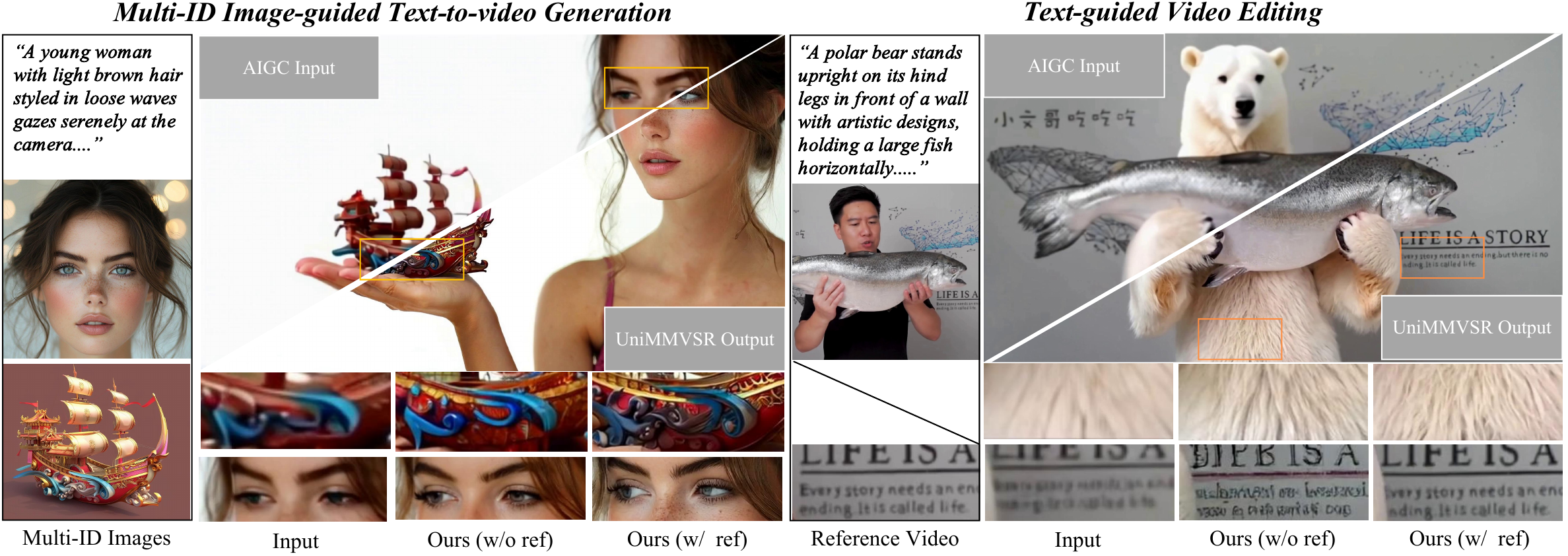}
\vspace{-1.5em}
\caption{\textbf{UniMMVSR is a unified framework that supports video super-resolution with multi-modal input conditions.} By cooperating with the low-resolution multi-modal generative model, the proposed cascaded framework can effectively extend the controllable video generation to ultra-high-resolution (e.g., 4K) with high visual quality and subject consistency.}
\label{fig:show}
\end{figure}

\begin{abstract}
Cascaded video super-resolution has emerged as a promising technique for decoupling the computational burden associated with generating high-resolution videos using large foundation models. Existing studies, however, are largely confined to text-to-video tasks and fail to leverage additional generative conditions beyond text, which are crucial for ensuring fidelity in multi-modal video generation. We address this limitation by presenting UniMMVSR, the first unified generative video super-resolution framework to incorporate hybrid-modal conditions, including text, images, and videos. We conduct a comprehensive exploration of condition injection strategies, training schemes, and data mixture techniques within a latent video diffusion model. A key challenge was designing distinct data construction and condition utilization methods to enable the model to precisely utilize all condition types, given their varied correlations with the target video. Our experiments demonstrate that UniMMVSR significantly outperforms existing methods, producing videos with superior detail and a higher degree of conformity to multi-modal conditions. We also validate the feasibility of combining UniMMVSR with a base model to achieve multi-modal guided generation of 4K videos—a feat previously unattainable with existing techniques.

\end{abstract}

%----------------------------------------

\section{Introduction}
\label{sec:introduction}

Video generation foundation models~\citep{bytedance2025seedance,wan2025wan,hu2025hunyuancustom} have made remarkable progress in synthesizing realistic videos, largely due to the scaling law of diffusion transformer architectures~\citep{peebles2023scalable}. Unfortunately, expanding model capacity typically incurs a significant computational burden, a challenge particularly pronounced for high-resolution video generation (e.g., 2K, 4K, 8K), a growing trend in future applications.
To resolve this dilemma, a stage-wise cascading paradigm, where a large-capacity base model generates a low-resolution video and subsequent lightweight super-resolution models synthesize the fine details, has emerged as a promising solution. Anyhow, existing research~\citep{ho2023imagenvideo,zhang2025flashvideo,bytedance2025seedance} on cascaded video super-resolution is limited to text-to-video task. A significant gap remains in understanding how super-resolution models can effectively use hybrid conditions—a crucial capability for maintaining generative fidelity in videos produced by multi-modal base models.

In this paper, we present the first unified latent diffusion framework for multi-modal video super-resolution, dubbed UniMMVSR. We focus on three common video generation tasks: text-to-video generation, multi-ID image-guided text-to-video generation, and text-guided video editing. For these tasks, our super-resolution model uses not only the low-resolution video but also text, ID images, and other videos as conditions. The main challenge is to integrate these diverse conditions into a single framework and to modulate these reference information in a compatible manner. This is crucial to ensure the model to use all conditions effectively, allowing it to generate vivid details that conform to the multi-modal guidance.

To achieve this, we conducted a thorough study on multi-modal condition injection, with a special focus on incorporating multiple ID images and reference videos. Our comparative analysis shows that token concatenation performs best among the baselines.
Recognizing that the low-resolution video from a base model might not perfectly align with the multi-modal conditions, we improved the robustness of UniMMVSR in two ways:
(i) We assign independent position embedding for condition tokens that are distinct from the noisy target video tokens. This encourages the model to use the references based on context and correlation even though their contents are pixel-aligned.
(ii) We developed a custom training data pipeline that simulates the generation characteristics of base models using the SDEdit technique~\citep{meng2021sdedit}.

Our experiments prove that UniMMVSR is superior to existing baselines, especially in its visual fidelity to multi-modal references. Our ablation studies further validate the effectiveness of our key designs, offering a clear view of the advantages of our method. We also show the benefits of our unified training framework: high-quality training data can transfer across sub-tasks, which reduces the burden of collecting high-quality data for complex-modal tasks.

Our contributions are summarized as follows:
\begin{itemize}
    \item We introduce UniMMVSR, the first multi-modal guided generative video super-resolution model built on a cascaded framework. Our model synthesizes vivid details while maintaining high fidelity to conditional references.
    \item We developed a unique SDEdit-based degradation pipeline to create synthetic training data for multi-modal video super-resolution. It enhances the model's robustness to discrepancies between low-resolution video inputs and multi-modal conditions.
    \item Our UniMMVSR framework demonstrates the ability to leverage high-quality training data across multiple tasks and can easily scale to ultra-high-resolution generation (e.g., 4K) with efficient computational overhead.
\end{itemize}

%----------------------------------------
\section{Related Works}
\label{sec:related_works}
%----------------------------------------

\subsection{Multi-modal Video Generation}

With the advancement of video generative model, recent work has increasingly focused on enhancing the controllability of generated videos.~\citep{huang2025conceptmaster,chen2025multi,yuan2025identity,he2024id,hu2024animate,lei2025animateanything,ma2024magic,wei2024dreamvideo,zhang2025magic} introduced reference images to improve subject consistency in the output video, while methods such as~\citep{chen2024follow,tu2025videoanydoor,mou2024revideo,ye2025stylemaster,liew2023magicedit} achieved mask-based or instruction-based video editing by incorporating referenced videos. Although these approaches demonstrate promising results for specific controllable tasks, they fail to generalize across diverse tasks, hindering the broader application of controllable video generation.

To establish a unified framework for controllable generation tasks, previous work~\citep{ding2022delta,ju2023humansd} introduced multiple adapter modules to independently incorporate different reference conditions. This approach yielded poor performance while resulting in significant waste of model parameters. Consequently, recent methods such as FullDiT~\citep{ju2025fulldit,tan2025ominictrl} leverage in-context condition mechanisms to flexibly combine multi-modal input condition signals through self-attention module, achieving multi-task controllable video generation in a unified framework.

However, the computational complexity of the self-attention module increases quadratically with the number of tokens, hindering the scalability of such methods to more tasks and higher resolutions. While FullDiT2~\citep{he2025fulldit2} optimizes computational overhead for reference conditions through kv cache, block skipping, and token selection techniques, it remains inapplicable to unified frameworks or high-resolution scenarios. To address this limitation, we propose the first unified cascaded framework for high-resolution multi-modal video generation. This approach effectively achieves controllable high-resolution video generation while faithfully preserving multiple input conditions.

\subsection{Video Super-resolution}

Most previous works~\citep{chan2022investigating,cao2021video,chan2021basicvsr,chan2022basicvsr++} primarily focus on synthetic or real-world data by designing compositional synthetic degradation pipelines to model the degraded videos. With the widespread applications of video generation models, later approaches shift towards AI-generated data. Due to limited generative capabilities, previous methods tend to generate over-smooth results. Motivated by recent advances in diffusion models, several diffusion-based video super-resolution (VSR) methods~\citep{wang2023lavie,zhou2024upscale,yang2024motion,he2024venhancer,li2025diffvsr,wang2025seedvr,wang2025seedvr2} have been proposed, which show impressive performance and generate realistic details.

However, limited by recent video super-resolution framework, existing models can only take text prompt and input video as conditions, which hinders their applications towards controllable video generation tasks. Although producing fine details, due to the randomness of diffusion sampling process, it inevitably reduces the fidelity of the input video to multi-modal references, which further diminishes the controllability of the generated results. In this paper, for the first time, we design a generative video super-resolution framework that unifies the input of hybrid-modal conditions, which improves the visual quality of the input video while ensuring its fidelity to multi-modal conditions.

%----------------------------------------

\section{Method}
\label{sec:method}
%----------------------------------------

\begin{figure}[t]
    \centering
    \includegraphics[width=1.0\linewidth]{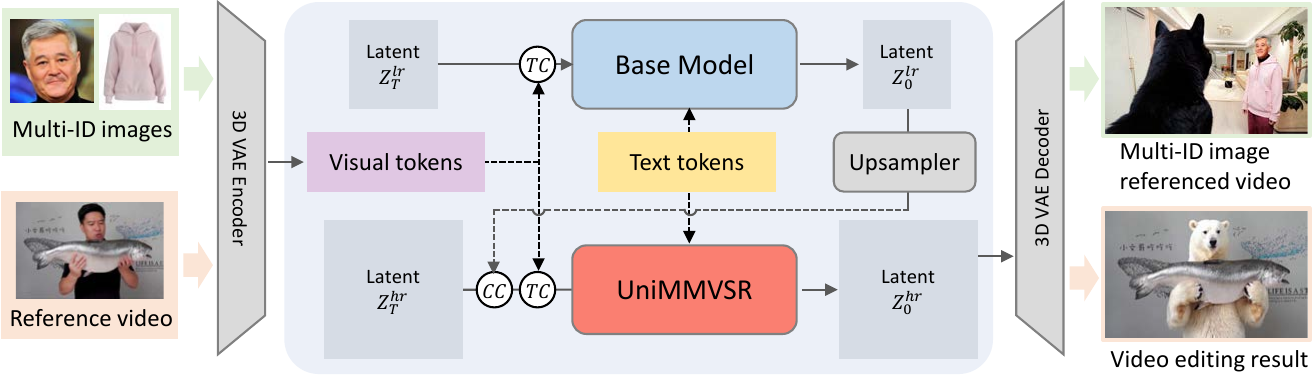}
    \vspace{-1.5em}
    \caption{Overview of UniMMVSR in the context of a cascaded generation framework. Upsampler denotes the sequential operations of VAE decoding, upscaling via bilinear interpolation, and VAE encoding. \textit{TC} and \textit{CC} denote token concatenation and channel concatenation respectively. Texts are encoded by text encoder and then injected via cross-attention layers, which are omit for simplicity.}
    \label{fig:architecture}
    \vspace{-1.0em}
\end{figure}

We aim to achieve generative video super-resolution for AI-generated videos under hybrid-modal conditions, which synthesizes rich, vivid details and maintains high fidelity to various conditional inputs. It works in the scenario that multi-modal base model first generates a low-resolution video, which our UniMMVSR model then upscales using the original high-resolution reference conditions if they're available. The overview flowchart is depicted in Fig.~\ref{fig:architecture}.
Specifically, we focus on three common video generation tasks: \textit{Text-to-video}, \textit{Text-to-video guided by multiple ID images}, and \textit{Text-guided video editing}. To accomplish this, our super-resolution model incorporates diverse inputs—including low-resolution video, text, multiple ID images, and reference videos—in a compatible manner. We have tackled this challenge by exploring a unified condition injection mechanism, a custom training data pipeline, and a tailored training strategy to ensure the model effectively utilizes all multi-modal conditions.

\subsection{Preliminaries}
Our model is built upon a pretrained large-scale text-to-video latent diffusion model. It first pretrains an autoencoder that converts a video $x$ into a low-dimensional latent $z$ with an encoder $\mathcal{E}$ and reconstructs it with a decoder $\mathcal{D}$. The core of this framework is a conditional diffusion transformer that operates in the compressed latent space. Detailed architecture is presented in Supp.~\ref{sec:supp-base-model}.

During training, given a LR-HR paired data $(z_{HR}, z_{LR})$ and multiple conditions $C$, isotropic gaussian noise is added to generate corresponding noise latent $z_t = (1-t) z_{HR} + t\epsilon$, where $\epsilon \in \mathcal{N}(0, I)$. With the formation of flow matching, it trains a network $\mu_\theta(z_t, t, C)$ to predict the velocity $v = z_{HR} - \epsilon$. Then, the network $\mu_\theta$ is optimized by minimizing the mean squared error loss $\mathcal{L}$ between the ground truth velocity and the model prediction:
\begin{equation}
    \label{eq:mse-loss}
    \mathcal{L} = \mathbb{E}_{z_{HR}, \epsilon, t, C}\Vert v - \mu_\theta(z_t, t, C)\Vert.
\end{equation}

During inference, it first samples noise $\epsilon \in \mathcal{N}(0, I)$, then denoises it by a pre-defined ODE solver with a discrete set of $N$ timesteps to generate clean latent $z_{HR}$. The final output $x_{HR}$ is obtained by projecting $z_{HR}$ to the pixel space using pre-trained decoder $\mathcal{D}$.

\subsection{Unified Conditioning Framework}

Our UniMMVSR model processes low-resolution input video with three types of reference conditions: text prompts, multiple ID images and reference videos. Since UniMMVSR is adapted from a pre-trained text-to-video model, it inherits the original text-prompt conditioning design. We mainly discuss the interaction strategy of conditional visual tokens and input video tokens.

\paragraph{Low-resolution video via channel concatenation.}
Since the low-resolution (LR) video and the target high-resolution (HR) video have pixel-aligned tempo-spatial correspondences, we use channel concatenation to directly incorporate the information of basic structure. To match their spatial sizes, we first upscale the LR video in pixel space and then encode it into a latent representation. This makes it ready for channel concatenation with the noisy HR latent\footnote{Note that, most VAE latent spaces do not support simple interpolation, which can cause significant structural distortion, as shown in ~\citep{xie2025simplegvr}}. During the inference phase, we first decode the LR latent generated by the base model. We then perform pixel interpolation to target resolution and finally encode it back into the latent space. This process ensures the LR latent has the same size as the noisy HR latent without corrupting the original information.

\paragraph{Visual references via token concatenation.}
We refer to multi-ID images and reference videos as visual references. Following the successful practice of recent in-context conditioning methods~\citep{tan2025ominictrl,ju2025fulldit}, we integrate the target video tokens with the conditional tokens using token concatenation. This strategy makes it feasible to incorporate general conditional modalities—whether they are spatially aligned or not—into the generation process, as they interact with the target video tokens in each layer's attention modules. Specifically, in each transformer block, noisy video tokens and visual references are processed in parallel through 2D self-attention, 2D cross-attention, and a feedforward network to preserve the alignment between the text and video modalities. For the 3D self-attention module, all tokens are treated as a single unified sequence and processed together, which ensures a bidirectional flow of information between the target video and visual reference tokens. Finally, all reference tokens are truncated from the transformer output to match the input shape.

\paragraph{Separated conditional RoPE.}
Before concatenation with the target video tokens, these conditional tokens are assigned with position encoding, where we adopt Rotary Position Embedding (RoPE)~\citep{su2024roformer}. For multi-ID images, it is natural to assign an individual range of RoPE that are distinct from that of target video tokens, since no direct spatial correspondence exists between them. For reference video, since the LR video generated by our base model is not perfectly pixel-aligned with it, we also assign a separate range of RoPE for thest conditional tokens, so as to encourage our model to utilize it based on context and correlation rather than direct copy-and-paste. Specifically, we assign indices $0$ to $f-1$ for noisy token, $n_i$ to $n_i+k_i$ for $i$-th reference tokens, where $f$ denotes the number of frames, $n_i$ and $k_i$ denotes the start index and length of $i$-th reference token.

\subsection{Degradation Pipeline}
Multi-modal conditional video generation model requires not only the vividness of generated content but also the conformity to provided conditions. Accordingly, our UniMMVSR model also needs to achieve high-quality details and maintains fidelity to the multi-modal conditions. Given a high-resolution video, it is of vital importance to design a degradation pipeline to process it into low-resolution video that simulates the degradation pattern of the base models' output. Specifically, the degradation pattern of multi-modal base model can be categorized into two scenarios:
(i) At low resolution, high-frequency details in training data are lost due to resize operations. The base model can only generate basic structures that align with the semantic content of the text prompt, which lacks fine details and textures.
(ii) For some challenging cases, the low-resolution output maintains a low fidelity to the visual references due to the sub-optimal controllability of the base model to harmonize text prompt and visual references. Therefore, the identity in LR video typically exhibits distortion in local structure and has low visual quality. 

To tackle these two scenarios, a custom degradation pipeline needs to be designed to simulate the artifacts and distortions generated by the base model. However, traditional degradation pipelines~\citep{wang2021real,chan2022investigating} constructed solely based on synthetic degradation factors (such as noise, blur, video compression, etc.) cannot be fully adapted to these scenarios since the resulting local structure of the LR video is strictly aligned with the HR video, failing to simulate the insufficient reference response in low-resolution output. To simulate the degradation scenario (ii), we recognized that \textbf{it is equivalent to the results obtained by base model using only text condition}. Therefore, based on the sdedit method~\citep{meng2021sdedit}, we constructed compatible high-frequency degradation features using inference result from the text-to-video base model, termed \textbf{SDEdit Degradation}. 

Specifically, we downsample HR video to a resolution directly achievable by the text-to-video base model. The resized video $x$ is encoded into latent space via a pre-trained 3D VAE encoder, and noise is added by the forward process of the diffusion model for $k$ steps, where the step value $k$ is randomly sampled from $[K_1, K_2]$ and $K_2$ denotes the maximum threshold to retain the main structure of the input video. Subsequently, we perform $k$ steps of denoising process on the noisy latent using the base model, decoding the result via the 3D VAE decoder to obtain the LR video. After sdedit degradation, we apply synthetic degradation factors to the output $x'$ to construct the final LR video. The degradation pipeline and samples are presented in Supp.~\ref{sec:supp-sdedit-results}.

\subsection{Training Strategy}
\label{sec:training-strategy}

\paragraph{Training Order.} To train a unified model with hybrid conditions, the training order of subtasks is essential due to the varied difficulty. Instead of generating by text prompt only, multi-ID image-guided text-to-video generation and text-guided video editing tasks tend to synthesize high-fidelity textures and details by utilizing visual conditions and text prompt together, thus resulting in faster convergence speed than text-to-video generation task as shown in Fig.~\ref{fig:supp-convergence-speed}. Thus, we perform a difficult-to-easy training strategy, aiming to learn the difficult task first, and then effectively adapt to easier tasks. 

Starting from a pre-trained text-to-video (T2V) model weight, we first train 21-frames text-to-video generation task independently in the first stage. In the second stage, we train 21-frames text-to-video generation and multi-ID image-guided text-to-video generation tasks together with a probability $0.6:0.4$, aiming to retain the ability to generate high-definition details from text. Next, all tasks are trained at 21 frames together with a probability $0.5:0.3:0.2$. Finally, we extend the frame length to 77 (5 seconds) while keeping the probability unchanged. 

\paragraph{Reference Augmentation.} For the multi-ID image-guided text-to-video generation task, most testing scenarios include cross-pair data, where the perspective, orientation, and position of the low-resolution output and ID images exhibit greater discrepancies than training datasets. For the text-guided video editing task, although the HR video and reference video are strictly pixel-aligned for non-editing area during training, the low-resolution output exhibits a certain degree of error compared with reference video. Since we only construct synthetic datasets for these two tasks (stated in Sec.~\ref{sec:supp-training-datasets}), directly utilizing synthetic reference conditions leads to a train-test gap, thereby compromising performance on the test set. To mitigate this issue, we design a reference augmentation technique to narrow this gap. Specifically, we apply several image-related transformations to simulate the cross-pair test scenarios of multi-ID image-guided text-to-video generation task. For the text-guided video editing task, we randomly shift the start frame of reference video, aiming to encourage the model to learn a more robust context-injection mechanism rather than directly copying the pixels from reference video.

%----------------------------------------
\begin{table}[t]
\vspace{-1.0em}
\centering
\footnotesize
\renewcommand{\arraystretch}{1.0}
\caption{Quantitative Evaluation of UniMMVSR on all three tasks. \textbf{Bold} and \underline{underlined} indicate the best and second-best results, respectively. $\uparrow$ indicates higher is better; $\downarrow$ indicates lower is better.}
\label{tab:quantitative}
\resizebox{\textwidth}{!}{
\begin{tabular}{@{}l ccc cc c ccc@{}}
\toprule
\multicolumn{10}{c}{\textbf{Text-to-video Generation}} \\
\midrule
\multicolumn{1}{c}{\multirow{2}{*}{\textbf{Method}}} & \multicolumn{4}{c}{\textbf{Visual Quality}} & \multicolumn{2}{c}{\textbf{Subject Consistency}} & \multicolumn{3}{c}{\textbf{Video Alignment}} \\
\cmidrule(lr){2-5} \cmidrule(lr){6-7} \cmidrule(lr){8-10}
& \textbf{MUSIQ$\uparrow$}& \textbf{CLIP-IQA$\uparrow$} & \textbf{QAlign$\uparrow$} & \textbf{DOVER$\uparrow$} & \textbf{CLIP-I$\uparrow$} & \textbf{DINO-I$\uparrow$} & \textbf{PSNR$\uparrow$} & \textbf{SSIM$\uparrow$} & \textbf{LPIPS$\downarrow$} \\
\midrule
Base 512$\times$512 & 30.996 & 0.246 & 3.741 & 0.594 & - & - & - & - & - \\
Base 1080P & 46.645 & 0.306 & 4.246 & 0.749 & - & - & - & - & - \\
VEnhancer & \textbf{57.171} & 0.367 & 4.214 & 0.733 & - & - & - & - & - \\
STAR & \underline{56.904} & 0.369 & 4.435 & 0.769 & - & - & - & - & - \\
SeedVR & 55.596 & \textbf{0.379} & 4.396 & \textbf{0.778} & - & - & - & - & - \\
Ours (single) & 56.146 & 0.366 & \textbf{4.535} & \underline{0.771} & - & - & - & - & - \\
\textbf{Ours (unified)} & 56.418 & \underline{0.371} & \underline{4.500} & \textbf{0.778} & - & - & - & - & - \\
\midrule
\multicolumn{10}{c}{\textbf{Text-guided Video Editing}} \\
\midrule
Base 512$\times$512 & 35.073 & 0.234 & 3.615 & 0.400 & - & - & 30.191 & 0.699 & 0.364 \\
Base 1080P & 53.616 & 0.383 & 4.247 & 0.634 & - & - & 29.383 & 0.582 & 0.358 \\
Ref Video & 54.249 & 0.365 & 4.131 & 0.571 & - & - & - & - & - \\
VEnhancer & 57.036 & 0.380 & 4.013 & 0.590 & - & - & 28.417 & 0.571 & 0.489 \\
STAR & 56.802 & \underline{0.397} & 4.264 & 0.608 & - & - & 29.421 & 0.631 & 0.397 \\
SeedVR & \underline{57.820} & 0.370 & 4.183 & \underline{0.635} & - & - & 29.535 & 0.597 & 0.413 \\
Ours (no ref) & \textbf{59.119} & \textbf{0.399} & 4.289 & \textbf{0.648} & - & - & 29.615 & 0.581 & 0.429 \\
Ours (single) & 53.388 & 0.348 & \underline{4.302} & 0.597 & - & - & \textbf{31.905} & \textbf{0.723} & \textbf{0.276} \\
\textbf{Ours (unified)} & 53.245 & 0.344 & \textbf{4.305} & 0.597 & - & - & \underline{31.556} & \underline{0.713} & \underline{0.282} \\
\midrule
\multicolumn{10}{c}{\textbf{Multi-ID Image-guided Text-to-video Generation}} \\
\midrule
Base 512$\times$512 & 29.314 & 0.255 & 3.149 & 0.433 & 0.692 & 0.538 & - & - & - \\
Base 1080P & 46.780 & 0.345 & 4.092 & 0.662 & 0.691 & 0.507 & - & - & - \\
VEnhancer & 60.656 & \textbf{0.469} & 4.149 & 0.707 & 0.671 & 0.533 & - & - & - \\
STAR & 58.810 & 0.449 & 4.282 & \textbf{0.763} & 0.696 & \underline{0.546} & - & - & - \\
SeedVR & 54.491 & 0.419 & 3.960 & 0.708 & 0.693 & 0.543 & - & - & - \\
Ours (no ref) & 60.947 & 0.445 & 4.385 & 0.742 & 0.693 & 0.543 & - & - & - \\
Ours (single) & \underline{61.357} & 0.446 & \underline{4.414} & 0.743 & \textbf{0.728} & \textbf{0.566} & - & - & - \\
\textbf{Ours (unified)} & \textbf{62.248} & \underline{0.465} & \textbf{4.428} & \underline{0.745} & \underline{0.726} & \textbf{0.566} & - & - & - \\
\bottomrule
\end{tabular}}
\vspace{-2.0em}
\end{table}

\section{Experiments}
\label{sec:experiments}

\subsection{Experimental Settings}
\begin{figure}[t]
    \vspace{-2.0em}
    \centering
    \includegraphics[width=1.0\linewidth]{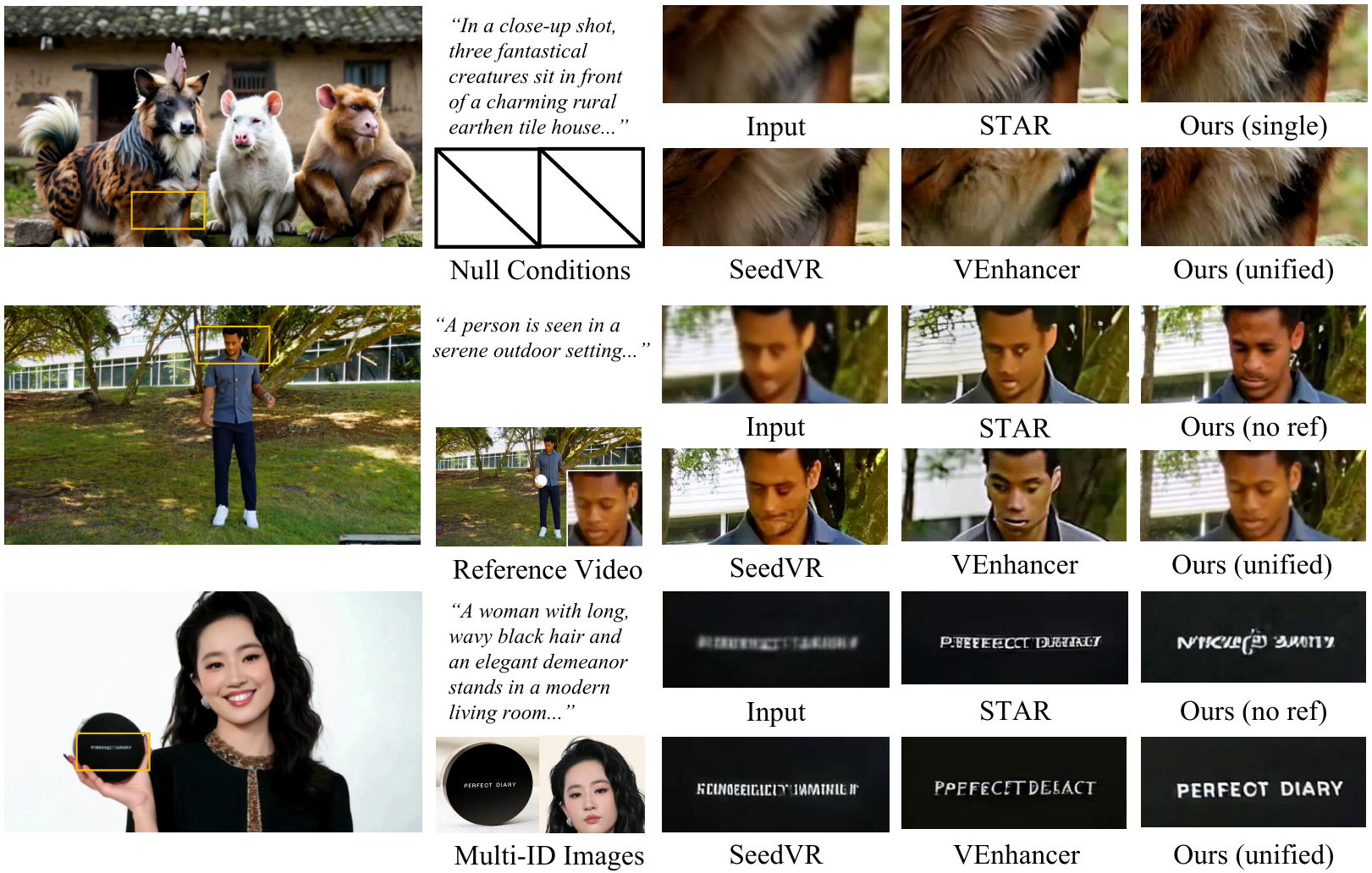}
    \vspace{-1.5em}
    \caption{Qualitative comparisons on \textit{text-to-video generation}, \textit{text-guided video editing} and \textit{multi-ID image-guided text-to-video generation} tasks from top to bottom. \textbf{(Zoom-in for best view)}}
    \label{fig:qualitative}
    \vspace{-2.0em}
\end{figure}

\subsubsection{Implementation Details} Our model is trained on NVIDIA H800 GPUs with a total batch size of 32. AdamW~\citep{loshchilov2017decoupled} is used as the optimizer with a learning rate of $10^{-4}$. The text prompt is randomly replaced by a null prompt with $10\%$ probability. To enhance the robustness of our model to different degradation scenarios, we utilize the noise augmentation technique by injecting noise into the input latent using a diffuse process. The noise timestep is randomly sampled from 200 to 600 to preserve the main structure. We have additionally encoded noise timestep as a micro condition for the model. We use a pretrained T2V model to provide initialization weight. During inference, we perform 50 PNDM~\citep{liu2022pseudo} sampling steps with independent classifier-free guidance as stated in Sec.~\ref{sec:supp-inference-technique}. The guidance scale $s_{txt}$ and $s_{ref}$ are set to $3.0$ and $1.0$ respectively, with reference guidance threshold $N_{ref}=15$. We have also used timestep shift~\citep{esser2024scaling} with shift value $1.0$. 

\subsubsection{Testing Settings}
\paragraph{Baseline Methods.} To evaluate the effect of our cascaded framework, we compare with the end-to-end results of our base model (both 512$\times$512 and 1080P). Since there is limited work on multi-modal VSR tasks, we also compare UniMMVSR with state-of-the-art VSR methods VEnhancer~\citep{he2024venhancer}, STAR~\citep{xie2025star} and SeedVR~\citep{wang2025seedvr,wang2025seedvr2}. 

\paragraph{Evaluation Metrics.} For visual quality, we conduct our evaluation of commonly used visual quality metrics MUSIQ~\citep{ke2021musiq}, CLIP-IQA~\citep{wang2023exploring}, Q-Align~\citep{wu2023q} and DOVER~\citep{wu2023exploring}. For multi-ID image-guided text-to-video generation task, we utilize DINO-I~\citep{caron2021emerging} and CLIP-I~\citep{radford2021learning} to assess the fidelity to multiple ID images. We additionally use PSNR, SSIM and LPIPS~\citep{zhang2018unreasonable} to evaluate alignment of non-editing area with the reference video for text-guided video editing task.

\subsection{Quantitative Comparison}
Quantitative comparisons are shown in Tab.~\ref{tab:quantitative}. Results show that although the UniMMVSR integrates multiple conditions, it still achieves state-of-the-art performance on controlling metrics compared with base model, previous VSR methods and our method without reference conditions, thereby validating the effectiveness of our method. For visual quality, UniMMVSR obtains the best QAlign$\&$DOVER scores on text-to-video generation task and MUSIQ$\&$QAlign scores on multi-ID image-guided text-to-video generation task, indicating its high perceptual quality. For text-guided video editing task, it is worth noting that our method maintains high pixel-level fidelity and structural similarity to the reference video for non-editing area, thus achieving similar metric values to the reference video. Even so, our approach remains competitive, achieving the best QAlign score. Furthermore, on multi-ID image-guided text-to-video generation task, our unified model exhibits high perceptual quality than our single-task model, indicating that complex-modal tasks can benefit from high-quality text-to-video data, effectively lowers the barrier to collect high-quality reference-video paired data. More comprehensive results can be seen in Supp.~\ref{sec:supp-quantitative-comparison}.

\begin{table}[t] % Using [ht] as provided, consider [!htbp] for more flexibility
\centering
\small % Use a smaller font for the table
\renewcommand{\arraystretch}{1.1} % Slightly reduce line spacing (same as previous)
\vspace{-3.0em}
\caption{Ablation Study of UniMMVSR components on the multi-ID image-guided text-to-video generation Task. We analyze the impact of each component by visual quality and controlling metrics.}
\label{tab:ablation}
\begin{adjustbox}{width=\textwidth,center} % Adjust table width to textwidth and center it
\begin{tabular}{@{}llcccccc@{}} % @{} removes padding at the table's edges
\toprule
\textbf{Ablation} & \textbf{Variant} & \textbf{MUSIQ$\uparrow$} & \textbf{CLIP-IQA$\uparrow$} & \textbf{QAlign$\uparrow$} & \textbf{DOVER$\uparrow$} & \textbf{CLIP-I$\uparrow$} & \textbf{DINO-I$\uparrow$} \\
\midrule
Ours & - & 62.248 & 0.465 & 4.428 & 0.745 & 0.726 & 0.566 \\
\midrule
\multirow{2}{*}{Architecture Design} & full channel-concat & 61.146 & 0.461 & 4.399 & 0.748 & 0.690 & 0.546 \\
 & full token-concat & 61.974 & 0.464 & 4.442 & 0.739 & 0.728 & 0.565 \\
\midrule
\multirow{2}{*}{Degradation Effect} & synthetic degradation only & 62.541 & 0.458 & 4.408 & 0.749 & 0.717 & 0.561 \\
 & sdedit degradation only & 59.697 & 0.437 & 4.357 & 0.726 & 0.730 & 0.564 \\
\midrule
\multirow{2}{*}{Training Order}& full training & 62.199 & 0.460 & 4.322 & 0.745 & 0.716 & 0.553 \\
 & easy-to-difficult & 61.706 & 0.445 & 4.326 & 0.736 & 0.717 & 0.556 \\
\bottomrule
\end{tabular}
\end{adjustbox} % Closes adjustbox
\end{table}

\subsection{Qualitative Comparison}
\label{sec:qualitative-comparison}
Fig.~\ref{fig:qualitative} shows visual results on all three tasks. For text-to-video generation, both our single-task and unified model effectively remove existing degradation patterns and generate fine details like the dog's fur, while other approaches produce blurred details. For text-guided video editing and multi-ID image-guided text-to-video generation, UniMMVSR successfully leverages ID images and reference videos to generate high-fidelity textures and details, such as the facial structure of the man and the words ``perfect diary'' on the box. More results can be found in Supp.~\ref{sec:supp-qualitative-comparison}.

\begin{figure}[t]
% \vspace{-1.0em}
\centering
\begin{minipage}[t]{0.49\textwidth}
\centering
\includegraphics[width=7cm]{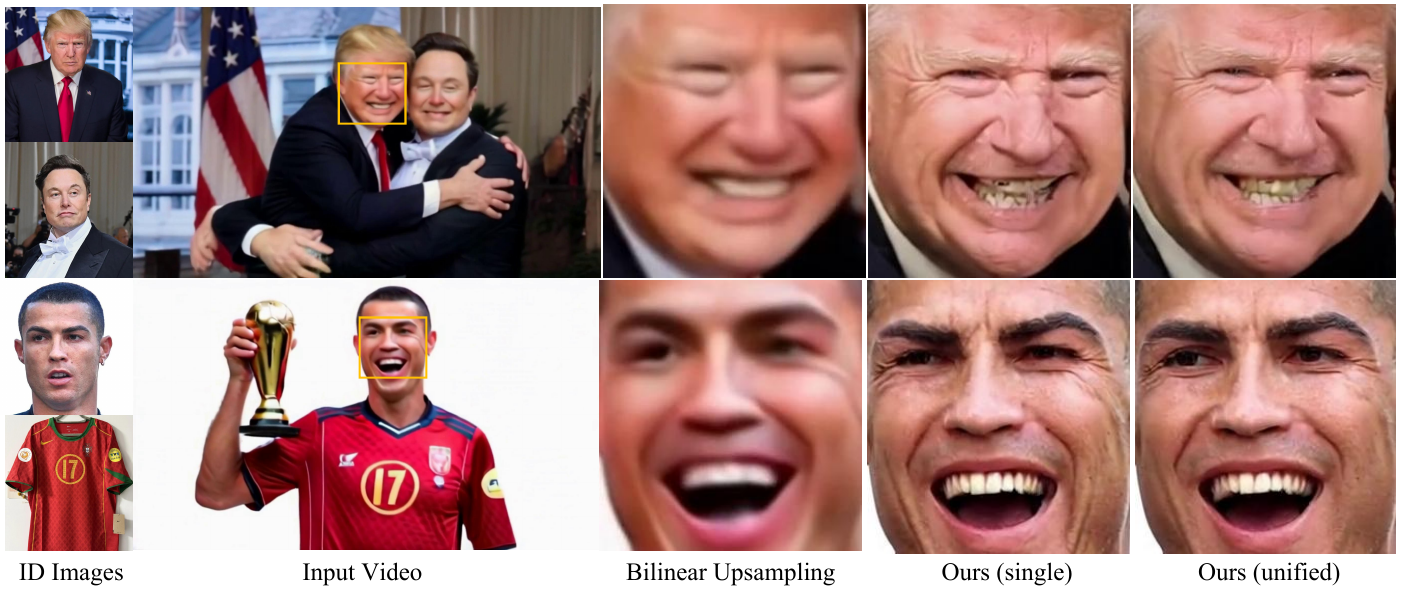}
\caption{Visual Comparisons of single-task and unified model. \textbf{Zoom-in for best view.}}
\label{fig:data-transfer}
\end{minipage}
\begin{minipage}[t]{0.49\textwidth}
\centering
\includegraphics[width=6cm]{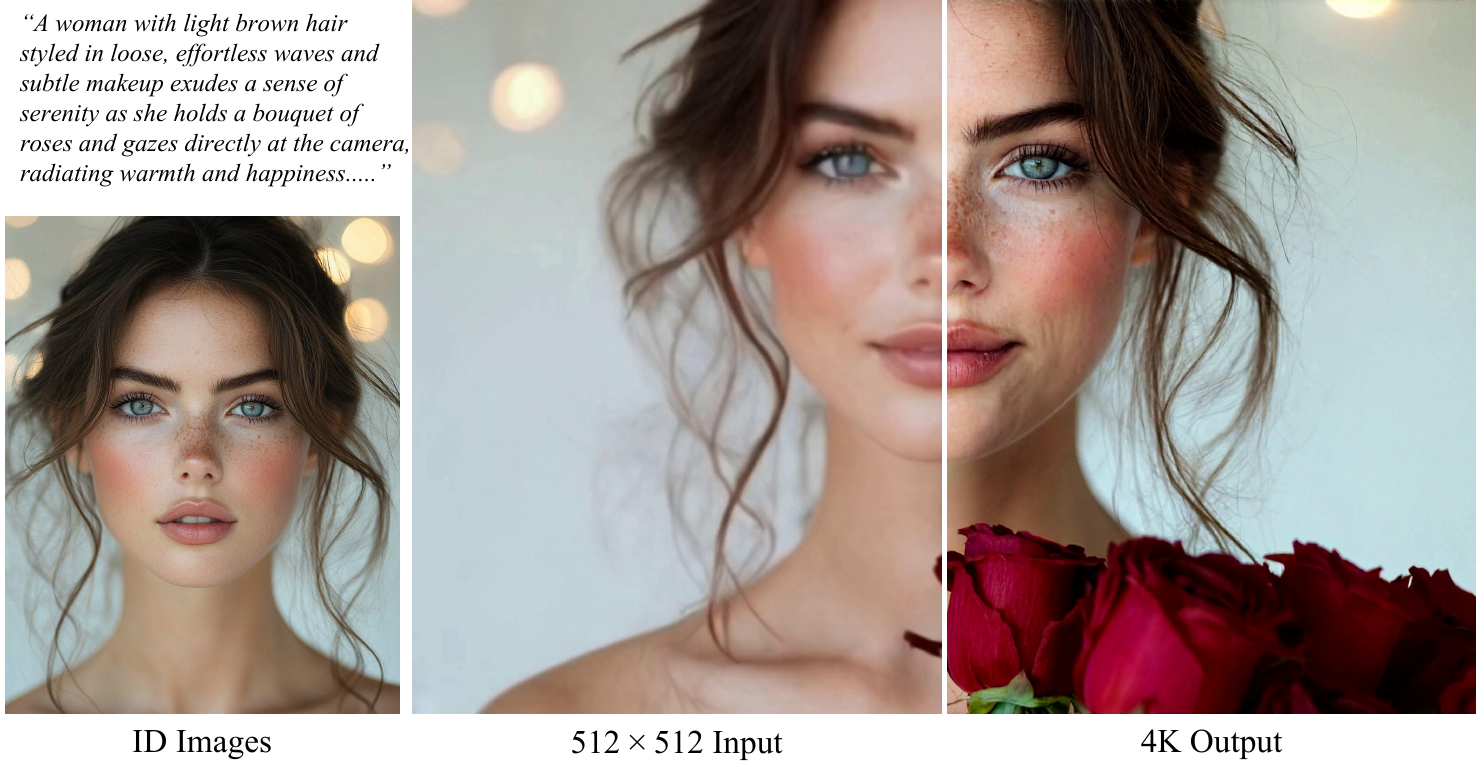}
\caption{Qualitative results of 4K multi-ID image-guided text-to-video generation.}
\label{fig:4k}
\end{minipage}
\vspace{-1.0em}
\end{figure}

\begin{figure}[t]
    \centering
    \includegraphics[width=1.0\linewidth]{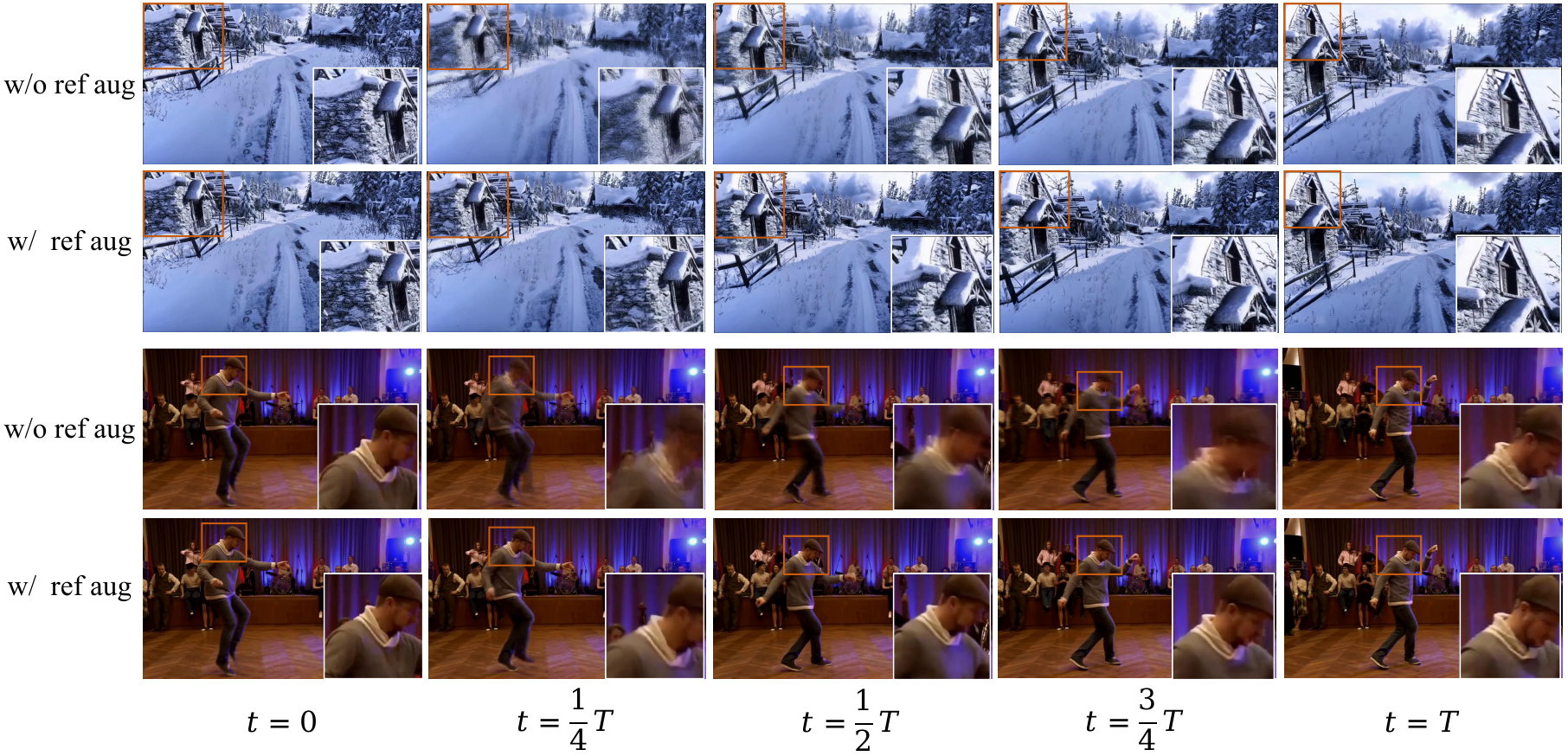}
    \vspace{-1.5em}
    \caption{Visual Comparisons of reference augmentation. \textbf{Zoom-in for best view.}}
    \label{fig:reference-augmentation}
    \vspace{-1.0em}
\end{figure}

\subsection{Ablation Study}
\label{sec:ablation-study}

Due to space limit, we provide qualitative evaluation in Supp.~\ref{sec:supp-ablation-study}.

\paragraph{Architecture Design.} In Tab.~\ref{tab:ablation}, we compare UniMMVSR with two architecture designs: full channel-concat and full token-concat. The former represents concatenating input video and reference tokens along channel dimension, while the latter represents concatenating along sequence dimension. As can be seen, full channel-concat method results in severe performance degradation in controlling metrics (0.690 vs 0.726 for CLIP-I and 0.546 vs 0.565 for DINO-I), which shows that it faces difficulties in reference injection. For full token-concat, while it achieves comparable performance, it results in nearly $2\times$ computational burden due to the quadratic computation complexity.

\paragraph{Degradation Effect.} Since our degradation pipeline comprises both synthetic and sdedit degradations, we perform ablation study to investigate the effectiveness of each component. As can be seen in Tab.~\ref{tab:ablation}, although only using synthetic degradation obtains similar visual quality metrics, it leads to poorer quality in controlling metrics, which confirms that sdedit degradation successfully simulates the degradation scenario of base model results. Next, to validate the necessity of traditional synthetic degradation pipeline, we use sdedit degradation only to construct LR data. While it achieves comparable controlling metrics, it shows a decline in all visual quality metrics, demonstrating the effectiveness of synthetic degradation in detail synthesis.

\paragraph{Training Order.} To form a unified model, we have implemented three different training strategies: difficult-to-easy, easy-to-difficult and full training. Specifically, difficult-to-easy means training in the order: text-to-video $\rightarrow$multi-ID image-guided$\rightarrow$video editing, whereas easy-to-difficult denotes the reverse order. Full training represents training all tasks together from the pre-trained T2V model weight. In Tab.~\ref{tab:ablation}, we demonstrate the validness of our proposed training strategy by comparing with other strategies above. By utilizing a difficult-to-easy training order, UniMMVSR successfully adapts to multiple tasks while maintaining the performance on previous tasks.

\subsection{Discussion}
\label{sec:exp-discussion}

\paragraph{Importance of Reference Augmentation.} Fig.~\ref{fig:reference-augmentation} shows visual comparisons on text-guided video editing task. As can be seen in the second frame in the upper part and the middle three frames in the lower part, training without reference augmentation tend to produce unstable structure, which results in temporal jitter in some frames. After using reference augmentation, UniMMVSR learns to preserve the basic structure of the input video and avoids direct replication of the reference video, which mitigates the conflict of the basic structure between reference video and LR video.

\paragraph{High-quality Data Transfer across Sub-tasks.} We perform an ablation study by training a single-task model on multi-ID image-guided text-to-video generation task without quality filtering. The model is compared with a unified model mix-trained on a high-quality text-to-video generation dataset. The results are shown in Fig.~\ref{fig:data-transfer}. As can be seen, the unified model generates more natural details such as teeth structure and facial expression, which demonstrates that the high resolution training data can transfer across sub-tasks.

\paragraph{Resolution Scaling Ability of Cascaded Model.} Due to the scarcity of ultra-high-resolution (UHR) reference-video paired dataset and quadratic computational complexity, it is difficult to directly train a high-resolution controllable video generative model. By decoupling the process as low-resolution basic structure generation and high-frequency detail synthesis, UniMMVSR successfully generates 4K videos under multi-modal guidance. As shown in Fig.~\ref{fig:4k} and Supp.~\ref{sec:supp-4k-results}, our method not only generates vivid details, but also preserves information in reference conditions.

%----------------------------------------

\section{Conclusion}
\label{sec:conclusion}

In this paper, we introduce UniMMVSR, the first multi-modal guided generative video super-resolution model built on a cascaded framework. By treating the visual references as a unified sequence and processing them via the 3D self-attention module, UniMMVSR effectively synthesizes vivid details while maintaining high fidelity to conditional references. To enhance the model's robustness to discrepancies
between low-resolution video inputs and multi-modal conditions, we develop a unique degradation pipeline based on sdedit method, which simulates the insufficient reference response in low-resolution output. Furthermore, we design a tailored training strategy to form a unified model, and demonstrate that high-quality training data can transfer across sub-tasks, which reduces
the burden of collecting high-quality data for complex-modal tasks. The proposed cascaded framework shows its resolution scaling ability, which achieves multi-modal guided 4K video generation for the first time.

%----------------------------------------

\newpage
\appendix
\section{Appendix}

The content in the appendix is categorized as follows:
\begin{itemize}
    \item Base Model.
    \item Training Datasets.
    \item Inference Technique.
    \item SDEdit Degradation.
    \item More Results.
        \begin{itemize}
            \item Training Convergence Speed.
            \item Quantitative Comparisons.
            \item Qualitative Comparisons.
            \item Ablation Study.
            \item 4K Results.
        \end{itemize}
\end{itemize}

\subsection{Base Model}
\label{sec:supp-base-model}
\begin{figure}[t]
    \centering
    \includegraphics[width=\linewidth]{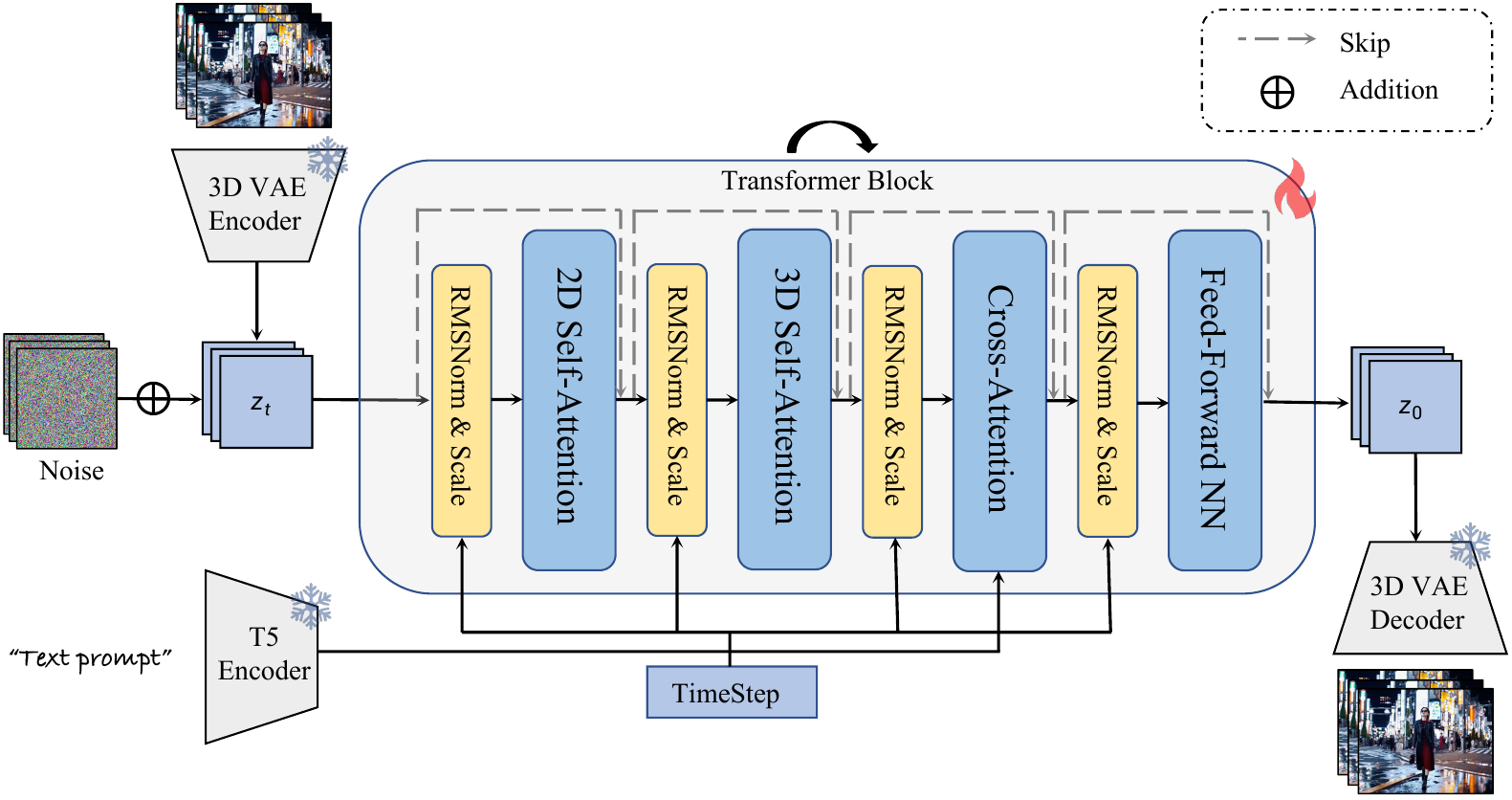}
    \caption{An overview of the architecture of our base model.}
    \label{fig:base-model}
\end{figure}

The architecture of our base model is shown in Fig.~\ref{fig:base-model}. Our UniMMVSR is built upon a pre-trained DiT-based video diffusion model, which comprises four main components: spatial self-attention (SSA), spatial cross-attention (SCA), temporal self-attention (TSA) and feed-forward network (FFN). Text prompts, time step and other micro conditions (aspect ratio, FPS, etc) are injected via the modulation mechanism~\citep{peebles2023scalable}. The pre-trained model is trained on 77-frames 512$\times$512 resolution high-quality video data with diverse aspect ratio using NaViT~\cite{dehghani2024patch}.

\subsection{Training Datasets}
\label{sec:supp-training-datasets}

\paragraph{Text-to-video Generation.} We train our model using 840K self-collected high-quality video-text pairs, with each clip processed to 5 seconds and 1080P resolution. The dataset is constructed by applying several IQA/VQA methods~\citep{wu2023q,wang2023exploring,ke2021musiq,wu2023exploring} to filter out low-quality data from 5M raw videos. The text prompts are all captioned using LLAVA captioner~\citep{liu2024visual}, and encoded by T5 text encoder~\citep{raffel2020exploring} with no more than 512 tokens.

\paragraph{Multi-ID Image-guided Text-to-video Generation.} Since portrait-related images dominate the application of multi-ID image-guided text-to-video generation task, we collect around 1.5M videos from open-sourced movies and television series. We then apply the same data filtering as text-to-video generation task to obtain 480K high-quality samples for training. We randomly select a non-overlapping frame from the video clip to extract the reference image. Finally, we apply Mask2former method~\citep{cheng2022masked} to identify and extract referenced images. 

\paragraph{Text-guided Video Editing.} Although inpainting-based datasets align better with test scenarios, model-generated reference videos naturally lack high-definition details required by our method. Thus, we follow the local-editing data pipeline~\citep{hu2024vivid} to preserve the high-frequency information in the non-editing area of the reference video, which results in 450K high-quality samples.

\subsection{Inference Technique}
\label{sec:supp-inference-technique}
UniMMVSR is trained on video data with either reference conditions or null conditions, and thus it can handle both scenarios. During inference, we apply independent classifier-free guidance (CFG) for each condition as:
\begin{equation}
\begin{aligned}
    \tilde{\epsilon}_\theta(z_t, t, c_{txt}, c_{ref}) &= \epsilon_\theta(z_t, t, c_{txt}, c_{ref}) \\
    &+ s_{txt}\cdot(\epsilon_\theta(z_t, t, c_{txt}, c_{ref}) - \epsilon_\theta(z_t, t, \phi_{txt}, c_{ref})) \\
    &+ s_{ref}\cdot(\epsilon_\theta(z_t, t, c_{txt}, c_{ref}) - \epsilon_\theta(z_t, t, c_{txt}, \phi_{ref})),
\end{aligned}
\end{equation}
where $c_{txt}$, $c_{ref}$ denote the condition of text prompt and reference, $\phi_{txt}$, $\phi_{ref}$ denote the corresponding null conditions and $s_{txt}$, $s_{ref}$ are the guidance scale. However, we find that simply increasing reference scale $s_{ref}$ leads to over-sharpen results and even generates artifacts. Since our goal is to modify the local structure of the input video based on reference conditions and generate high-frequency details, we introduce reference guidance threshold (RGT) technique to only utilize reference condition for first $N_{ref}$ steps as below:
\begin{equation}
    \tilde{s}_{ref} = \left\{
    \begin{aligned}
        s_{ref} & , & n < N_{ref} \\
        0 & , & n \geq N_{ref}
    \end{aligned}
    \right.
\end{equation}

We have compared the results of different inference settings in Fig.~\ref{fig:supp-reference-guidance-threshold}. As shown below, directly increasing reference guidance scale leads to over-sharpen details and even artifacts. By utilizing the proposed RGT technique ($s_{ref}=1.0\& N_{ref}=15$), UniMMVSR enhances the guidance of the reference conditions, further strengthening the generalization on the cross-pair test set.

\begin{figure}[t]
    \centering
    \includegraphics[width=1.0\linewidth]{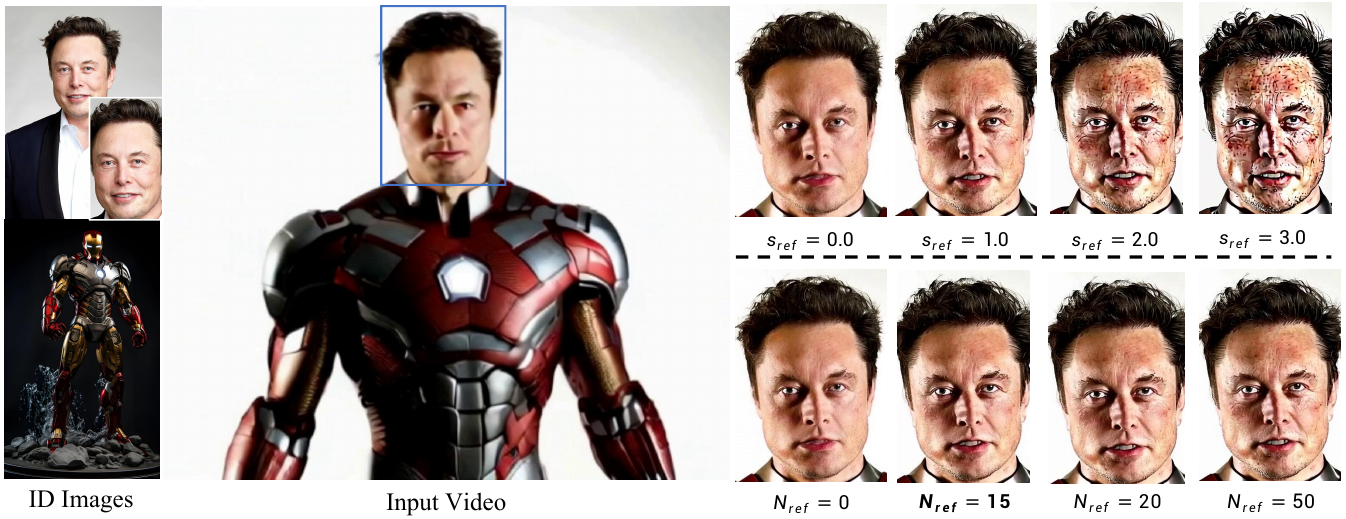}
    \caption{Qualitative comparisons of different inference settings. The text prompt is omitted.}
    \label{fig:supp-reference-guidance-threshold}
\end{figure}

\subsection{SDEdit Degradation}
\label{sec:supp-sdedit-results}
\begin{figure}[t]
    \centering
    \includegraphics[width=1.0\linewidth]{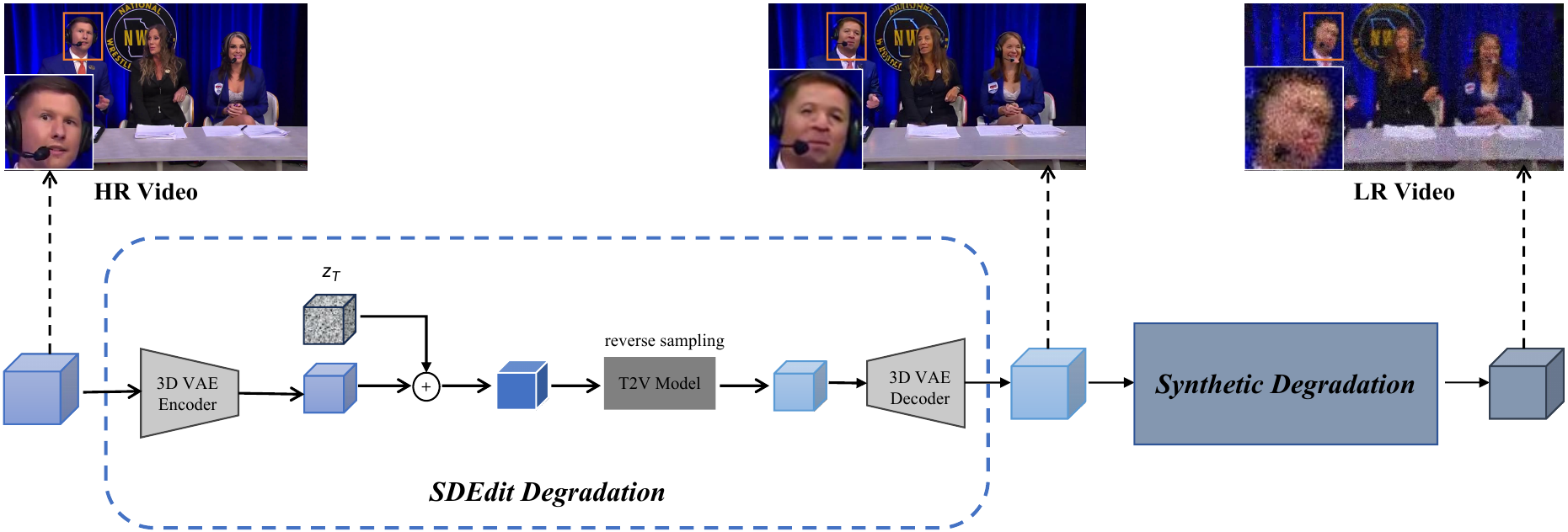}
    \caption{Degradation pipeline for UniMMVSR.}
    \label{fig:supp-degradation}
\end{figure}

\begin{figure}[t]
    \centering
    \includegraphics[width=1.0\linewidth]{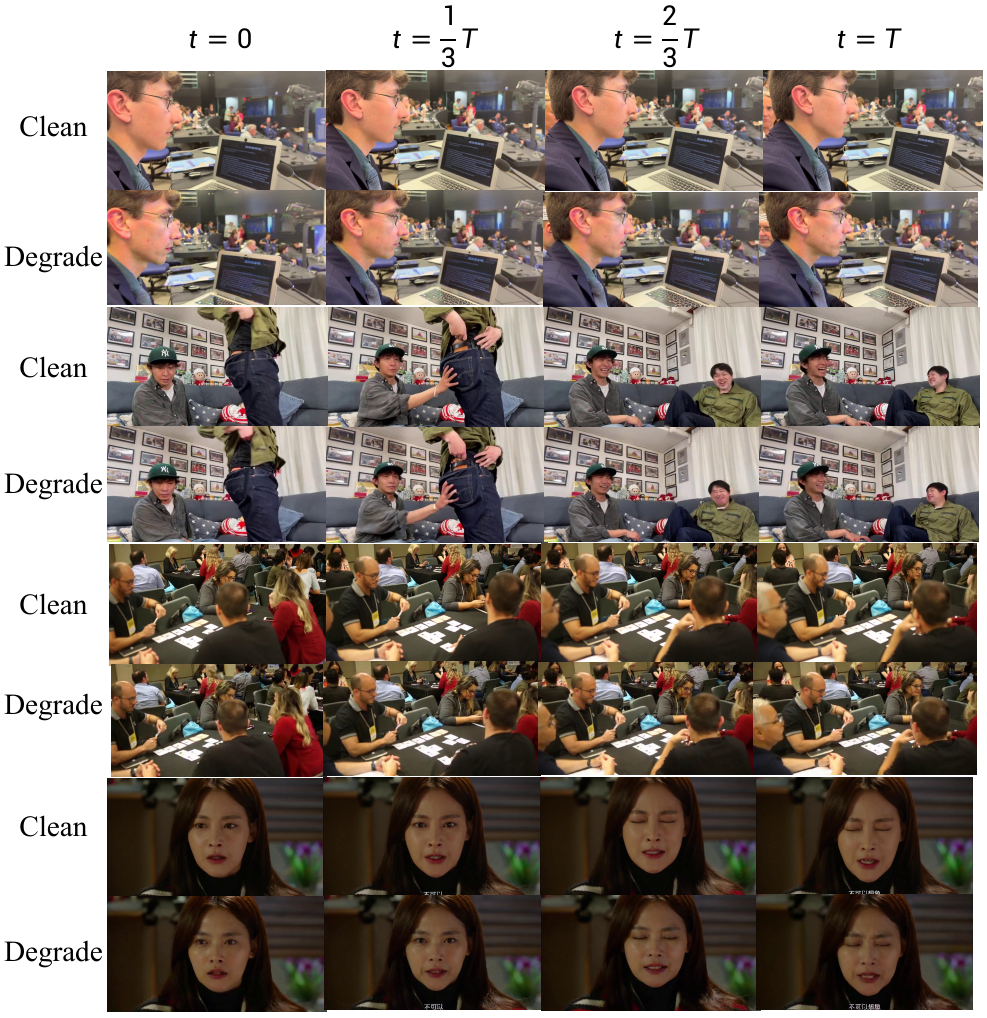}
    \caption{Samples of light sdedit degradation.}
    \label{fig:supp-sdedit-light}
\end{figure}

\begin{figure}[t]
    \centering
    \includegraphics[width=1.0\linewidth]{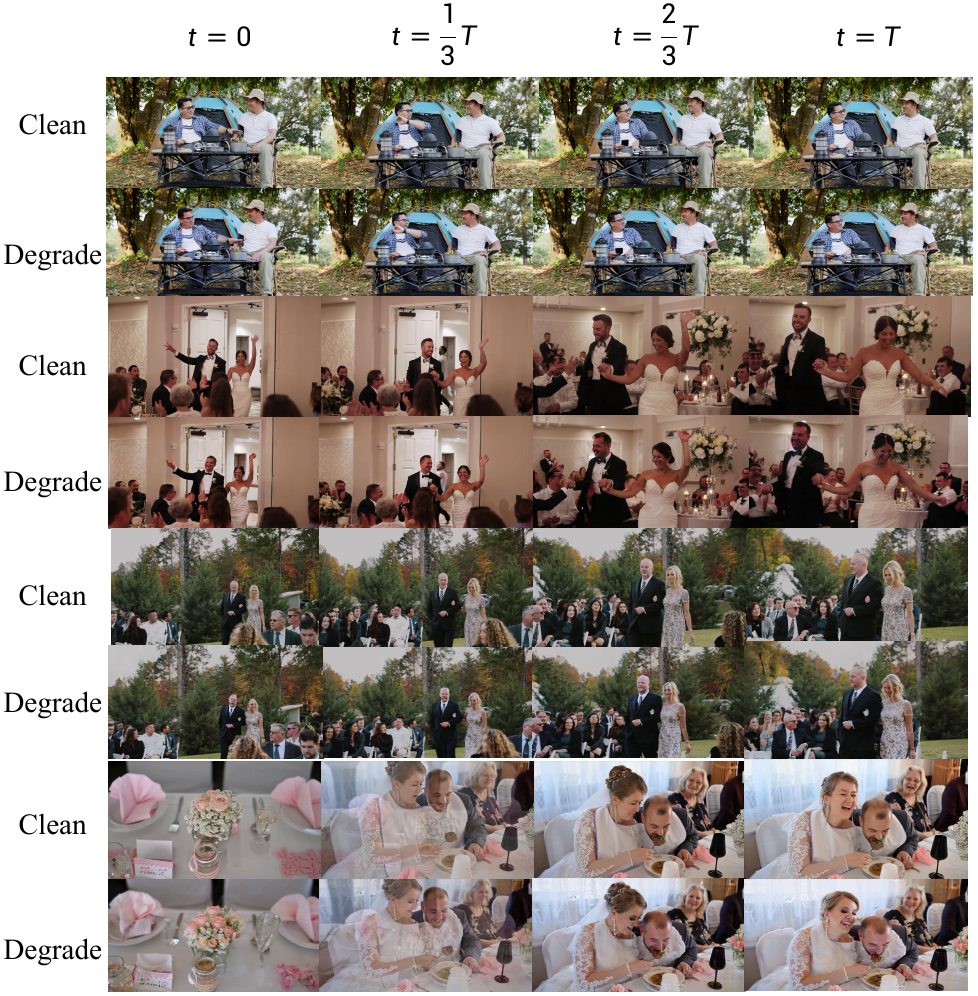}
    \caption{Samples of heavy sdedit degradation.}
    \label{fig:supp-sdedit-heavy}
\end{figure}

The degradation pipeline is shown in Fig.~\ref{fig:supp-degradation}. We first perform sdedit degradation to modify the local structure of HR video using the sdedit method by our text-to-video base model. Afterwards, we apply traditional synthetic degradation to introduce high-frequency degradation pattern. Light and heavy sdedit degradation samples are shown in Fig.~\ref{fig:supp-sdedit-light} and ~\ref{fig:supp-sdedit-heavy} respectively.

\subsection{More Results}

\subsubsection{Training Convergence Speed}
\label{sec:supp-training-convergence}
\begin{figure}
    \centering
    \includegraphics[width=1.0\linewidth]{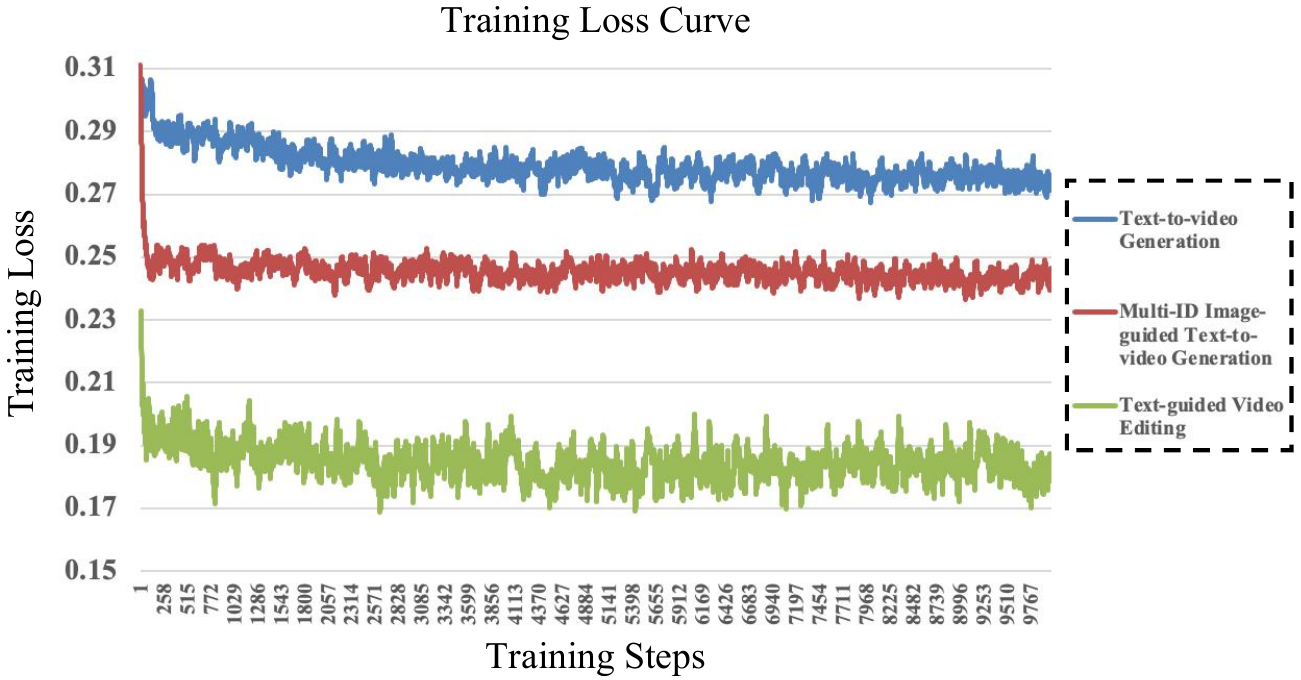}
    \caption{Training loss curve of all three tasks.}
    \label{fig:supp-convergence-speed}
\end{figure}

We have shown the training loss curve of single-task model on text-to-video generation, multi-ID image-guided text-to-video generation and text-guided video editing tasks in Fig.~\ref{fig:supp-convergence-speed}. As can be seen, text-guided video editing task tends to converge faster at a lower loss value $0.18$, while text-to-video generation task converges slowest, at around $3k$ steps.

\subsubsection{Quantitative Comparisons}
\label{sec:supp-quantitative-comparison}

Full quantitative comparisons of text-to-video generation, multi-ID image-guided text-to-video generation and text-guided video editing tasks are shown in Tab.~\ref{tab:supp-quantitative-original}, ~\ref{tab:supp-quantitative-id} and ~\ref{tab:supp-quantitative-editing} respectively.

\begin{table}[t]
\centering
\footnotesize
\renewcommand{\arraystretch}{1.0}
\caption{Quantitative comparison of text-to-video generation task. \textbf{Bold} and \underline{underlined} indicate the best and second-best results, respectively. $\uparrow$ indicates higher is better; $\downarrow$ indicates lower is better.}
\label{tab:supp-quantitative-original}
\resizebox{\textwidth}{!}{
\begin{tabular}{@{}l ccc cc c ccc@{}}
\toprule
\multicolumn{10}{c}{\textbf{Text-to-video Generation}} \\
\midrule
\multicolumn{1}{c}{\multirow{2}{*}{\textbf{Method}}} & \multicolumn{4}{c}{\textbf{Visual Quality}} & \multicolumn{2}{c}{\textbf{Subject Consistency}} & \multicolumn{3}{c}{\textbf{Video Alignment}} \\
\cmidrule(lr){2-5} \cmidrule(lr){6-7} \cmidrule(lr){8-10}
& \textbf{MUSIQ$\uparrow$}& \textbf{CLIP-IQA$\uparrow$} & \textbf{QAlign$\uparrow$} & \textbf{DOVER$\uparrow$} & \textbf{CLIP-I$\uparrow$} & \textbf{DINO-I$\uparrow$} & \textbf{PSNR$\uparrow$} & \textbf{SSIM$\uparrow$} & \textbf{LPIPS$\downarrow$} \\
\midrule
Base 512$\times$512 & 30.996 & 0.246 & 3.741 & 0.594 & - & - & - & - & - \\
Base 1080P & 46.645 & 0.306 & 4.246 & 0.749 & - & - & - & - & - \\
VEnhancer-v1 & \textbf{57.171} & 0.367 & 4.214 & 0.733 & - & - & - & - & - \\
VEnhancer-v2 & 44.603 & 0.364 & 4.091 & 0.693 & - & - & - & - & - \\
STAR-light & \underline{56.904} & 0.369 & 4.435 & 0.769 & - & - & - & - & - \\
STAR-heavy & 55.294 & 0.361 & \textbf{4.579} & \textbf{0.795} & - & - & - & - & - \\
SeedVR-7B & 55.596 & \textbf{0.379} & 4.396 & 0.778 & - & - & - & - & - \\
SeedVR-3B & 54.310 & \underline{0.375} & 4.281 & 0.762 & - & - & - & - & - \\
SeedVR2-7B & 48.490 & 0.310 & 4.292 & 0.763 & - & - & - & - & - \\
SeedVR2-7B-sharp & 47.013 & 0.300 & 4.234 & 0.753 & - & - & - & - & - \\
SeedVR2-3B & 50.604 & 0.328 & 4.318 & \underline{0.783} & - & - & - & - & - \\
Ours (single) & 56.146 & 0.366 & \underline{4.535} & 0.771 & - & - & - & - & - \\
\textbf{Ours (unified)} & 56.418 & 0.371 & 4.500 & 0.778 & - & - & - & - & - \\
\bottomrule
\end{tabular}}
\end{table}

\begin{table}[t]
\centering
\footnotesize
\renewcommand{\arraystretch}{1.0}
\caption{Quantitative comparison of multi-ID image-guided text-to-video generation task. \textbf{Bold} and \underline{underlined} indicate the best and second-best results, respectively. $\uparrow$ indicates higher is better; $\downarrow$ indicates lower is better.}
\label{tab:supp-quantitative-id}
\resizebox{\textwidth}{!}{
\begin{tabular}{@{}l ccc cc c ccc@{}}
\toprule
\multicolumn{10}{c}{\textbf{Multi-ID Image-guided Text-to-video Generation}} \\
\midrule
\multicolumn{1}{c}{\multirow{2}{*}{\textbf{Method}}} & \multicolumn{4}{c}{\textbf{Visual Quality}} & \multicolumn{2}{c}{\textbf{Subject Consistency}} & \multicolumn{3}{c}{\textbf{Video Alignment}} \\
\cmidrule(lr){2-5} \cmidrule(lr){6-7} \cmidrule(lr){8-10}
& \textbf{MUSIQ$\uparrow$}& \textbf{CLIP-IQA$\uparrow$} & \textbf{QAlign$\uparrow$} & \textbf{DOVER$\uparrow$} & \textbf{CLIP-I$\uparrow$} & \textbf{DINO-I$\uparrow$} & \textbf{PSNR$\uparrow$} & \textbf{SSIM$\uparrow$} & \textbf{LPIPS$\downarrow$} \\
\midrule
Base 512$\times$512 & 29.314 & 0.255 & 3.149 & 0.433 & 0.692 & 0.538 & - & - & - \\
Base 1080P & 46.780 & 0.345 & 4.092 & 0.662 & 0.691 & 0.507 & - & - & - \\
VEnhancer-v1 & 60.656 & \textbf{0.469} & 4.149 & 0.707 & 0.671 & 0.533 & - & - & - \\
VEnhancer-v2 & 43.776 & 0.422 & 3.860 & 0.628 & 0.690 & 0.538 & - & - & - \\
STAR-light & 58.810 & 0.449 & 4.282 & \textbf{0.763} & 0.696 & 0.546 & - & - & - \\
STAR-heavy & 54.446 & 0.399 & 4.223 & 0.721 & 0.695 & \underline{0.547} & - & - & - \\
SeedVR-7B & 54.491 & 0.419 & 3.960 & 0.708 & 0.693 & 0.543 & - & - & - \\
SeedVR-3B & 53.943 & 0.416 & 3.845 & 0.689 & 0.696 & 0.544 & - & - & - \\
SeedVR2-7B & 49.220 & 0.344 & 3.814 & 0.664 & 0.689 & 0.543 & - & - & - \\
SeedVR2-7B-sharp & 46.718 & 0.332 & 3.751 & 0.639 & 0.691 & 0.545 & - & - & - \\
SeedVR2-3B & 51.169 & 0.368 & 3.850 & 0.690 & 0.694 & 0.544 & - & - & - \\
Ours (no ref) & 60.947 & 0.445 & 4.385 & 0.742 & 0.693 & 0.543 & - & - & - \\
Ours (single) & \underline{61.357} & 0.446 & \underline{4.414} & 0.743 & \textbf{0.728} & \textbf{0.566} & - & - & - \\
\textbf{Ours (unified)} & \textbf{62.248} & \underline{0.465} & \textbf{4.428} & \underline{0.745} & \underline{0.726} & \textbf{0.566} & - & - & - \\
\bottomrule
\end{tabular}}
\end{table}

\begin{table}[t]
\centering
\footnotesize
\renewcommand{\arraystretch}{1.0}
\caption{Quantitative comparison of text-guided video editing task. \textbf{Bold} and \underline{underlined} indicate the best and second-best results, respectively. $\uparrow$ indicates higher is better; $\downarrow$ indicates lower is better.}
\label{tab:supp-quantitative-editing}
\resizebox{\textwidth}{!}{
\begin{tabular}{@{}l ccc cc c ccc@{}}
\toprule
\multicolumn{10}{c}{\textbf{Text-guided Video Editing}} \\
\midrule
\multicolumn{1}{c}{\multirow{2}{*}{\textbf{Method}}} & \multicolumn{4}{c}{\textbf{Visual Quality}} & \multicolumn{2}{c}{\textbf{Subject Consistency}} & \multicolumn{3}{c}{\textbf{Video Alignment}} \\
\cmidrule(lr){2-5} \cmidrule(lr){6-7} \cmidrule(lr){8-10}
& \textbf{MUSIQ$\uparrow$}& \textbf{CLIP-IQA$\uparrow$} & \textbf{QAlign$\uparrow$} & \textbf{DOVER$\uparrow$} & \textbf{CLIP-I$\uparrow$} & \textbf{DINO-I$\uparrow$} & \textbf{PSNR$\uparrow$} & \textbf{SSIM$\uparrow$} & \textbf{LPIPS$\downarrow$} \\
\midrule
Base 512$\times$512 & 35.073 & 0.234 & 3.615 & 0.400 & - & - & 30.191 & 0.699 & 0.364 \\
Base 1080P & 53.616 & 0.383 & 4.247 & 0.634 & - & - & 29.383 & 0.582 & 0.358 \\
Ref Video & 54.249 & 0.365 & 4.131 & 0.571 & - & - & - & - & - \\
VEnhancer-v1 & 57.036 & 0.380 & 4.013 & 0.590 & - & - & 28.417 & 0.571 & 0.489 \\
VEnhancer-v2 & 48.084 & 0.353 & 3.959 & 0.557 & - & - & 28.712 & 0.628 & 0.410 \\
STAR-light & 56.802 & \underline{0.397} & 4.264 & 0.608 & - & - & 29.421 & 0.631 & 0.397 \\
STAR-heavy & 56.207 & 0.378 & 4.259 & 0.599 & - & - & 29.442 & 0.648 & 0.371 \\
SeedVR-7B & \underline{57.820} & 0.370 & 4.183 & \underline{0.635} & - & - & 29.535 & 0.597 & 0.413 \\
SeedVR-3B & 55.326 & 0.360 & 4.048 & 0.628 & - & - & 29.338 & 0.588 & 0.416 \\
SeedVR2-7B & 54.046 & 0.361 & 4.087 & 0.610 & - & - & 29.600 & 0.614 & 0.367 \\
SeedVR2-7B-sharp & 52.723 & 0.359 & 4.010 & 0.579 & - & - & 29.563 & 0.619 & 0.362 \\
SeedVR2-3B & 54.310 & 0.355 & 4.099 & 0.597 & - & - & 29.326 & 0.593 & 0.382 \\
Ours (no ref) & \textbf{59.119} & \textbf{0.399} & 4.289 & \textbf{0.648} & - & - & 29.615 & 0.581 & 0.429 \\
Ours (single) & 53.388 & 0.348 & \underline{4.302} & 0.597 & - & - & \textbf{31.905} & \textbf{0.723} & \textbf{0.276} \\
\textbf{Ours (unified)} & 53.245 & 0.344 & \textbf{4.305} & 0.597 & - & - & \underline{31.556} & \underline{0.713} & \underline{0.282} \\
\bottomrule
\end{tabular}}
\end{table}

\subsubsection{Qualitative Comparisons}
\label{sec:supp-qualitative-comparison}
\begin{figure}[t]
    \centering
    \includegraphics[width=1.0\linewidth]{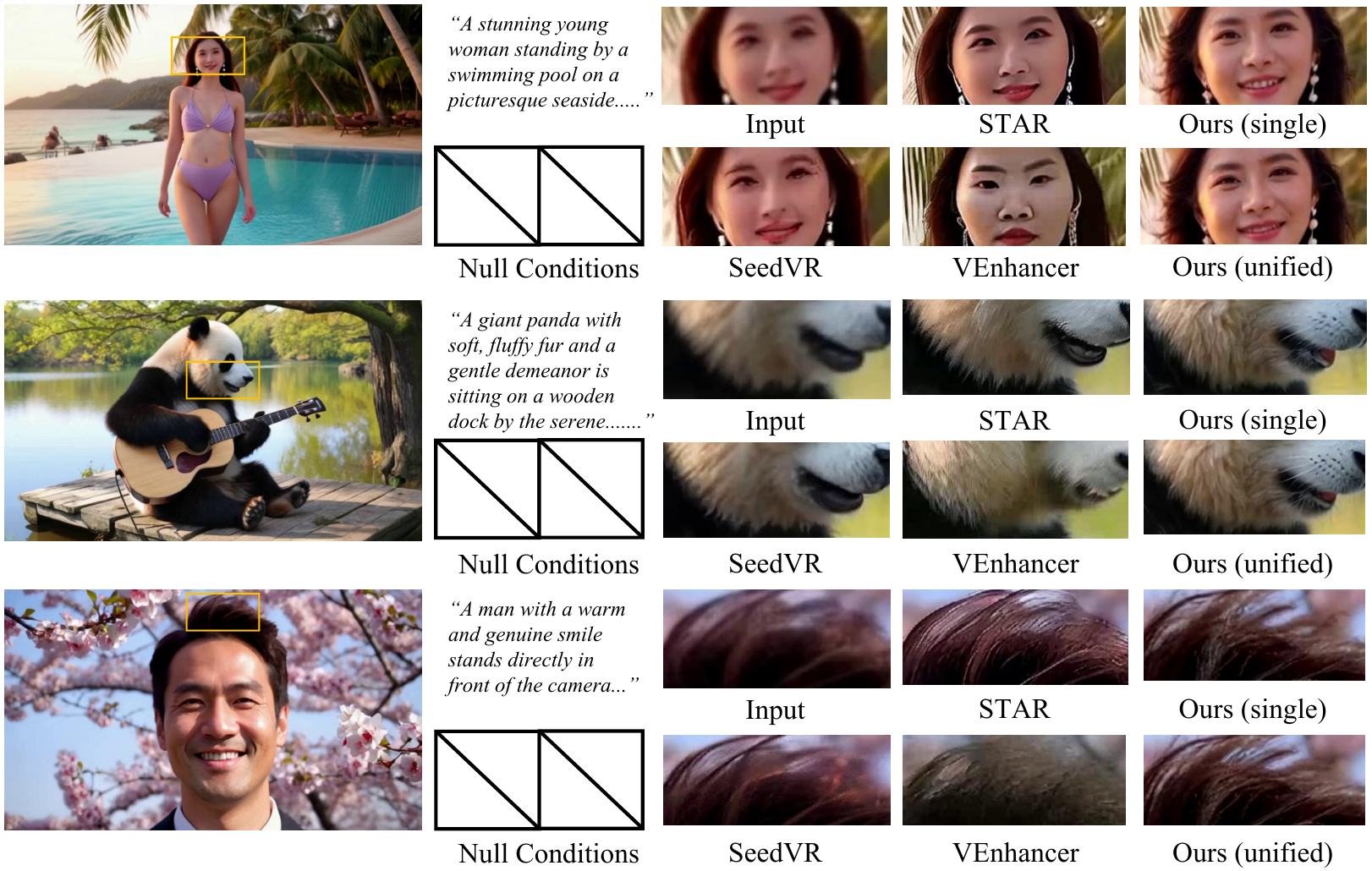}
    \caption{Qualitative comparisons on text-to-video generation task.}
    \label{fig:supp-qualitative-original}
\end{figure}

\begin{figure}[t]
    \centering
    \includegraphics[width=1.0\linewidth]{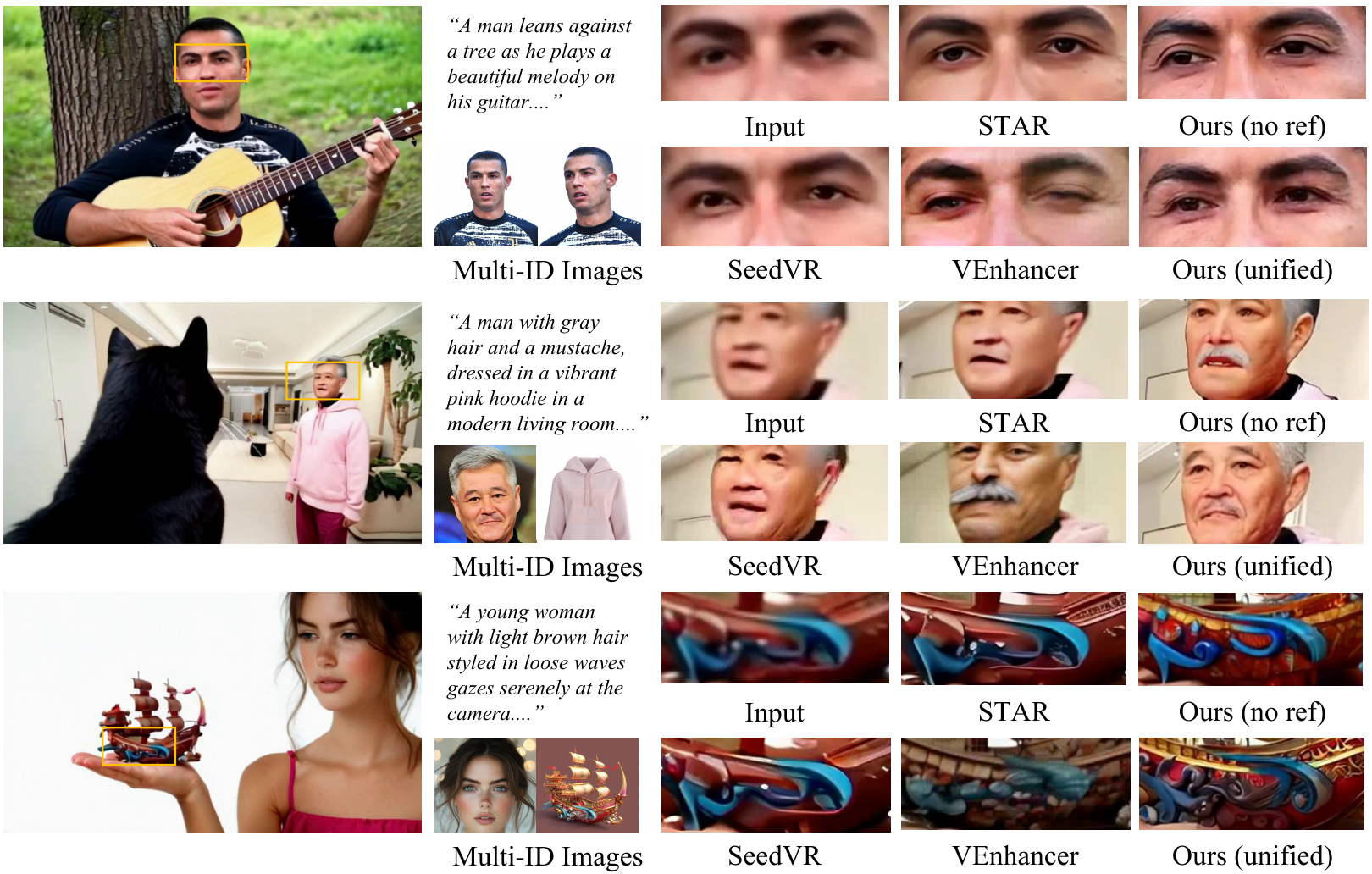}
    \caption{Qualitative comparisons on multi-ID image-guided text-to-video generation task.}
    \label{fig:supp-qualitative-id}
\end{figure}

\begin{figure}[t]
    \centering
    \includegraphics[width=1.0\linewidth]{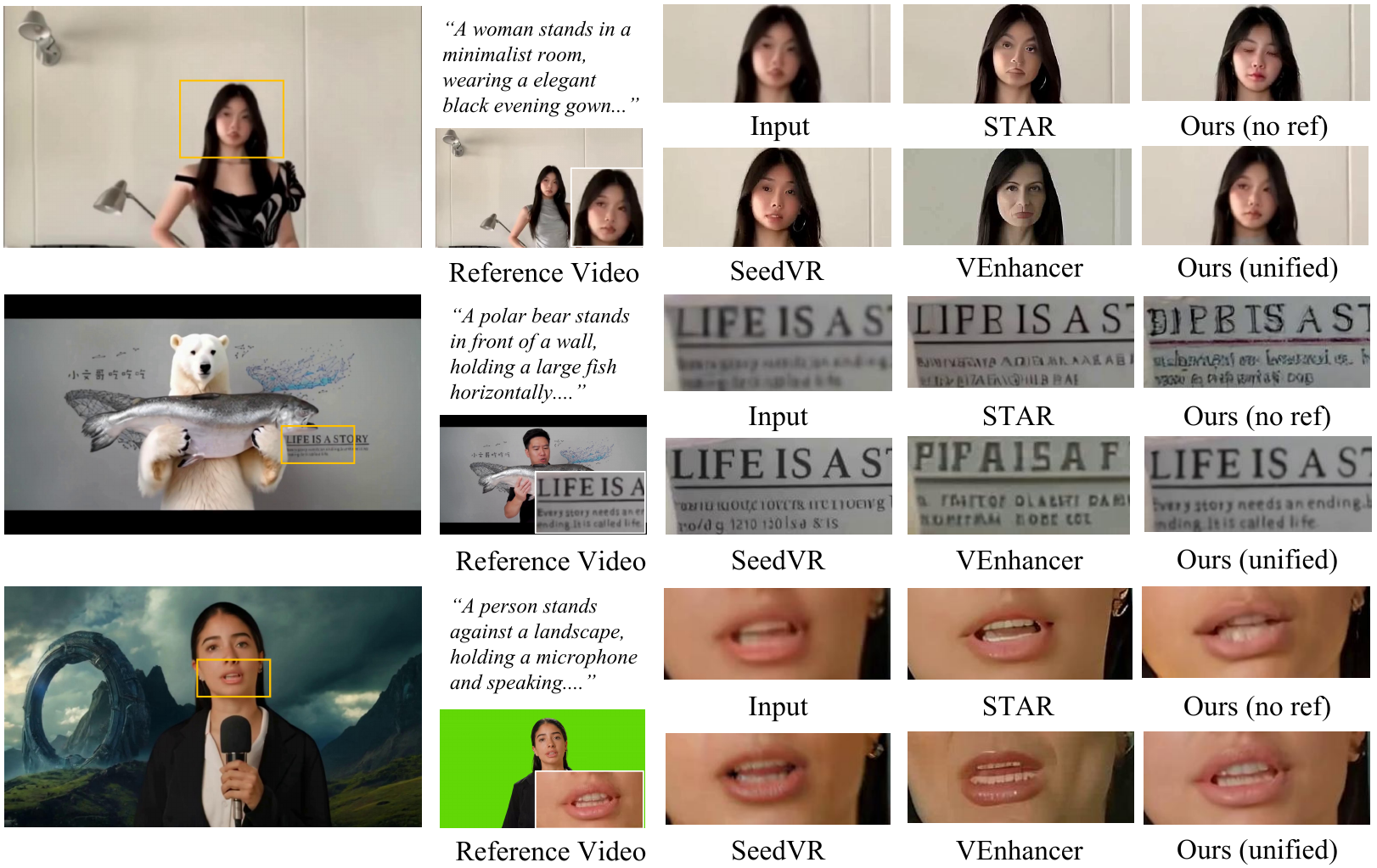}
    \caption{Qualitative comparisons on text-guided video editing task.}
    \label{fig:supp-qualitative-editing}
\end{figure}

Additional qualitative comparisons of text-to-video generation, multi-ID image-guided text-to-video generation and text-guided video editing tasks are presented in Fig.~\ref{fig:supp-qualitative-original}, ~\ref{fig:supp-qualitative-id} and ~\ref{fig:supp-qualitative-editing} respectively.

\subsubsection{Ablation Study}
\label{sec:supp-ablation-study}

\begin{figure}
    \centering
    \includegraphics[width=1.0\linewidth]{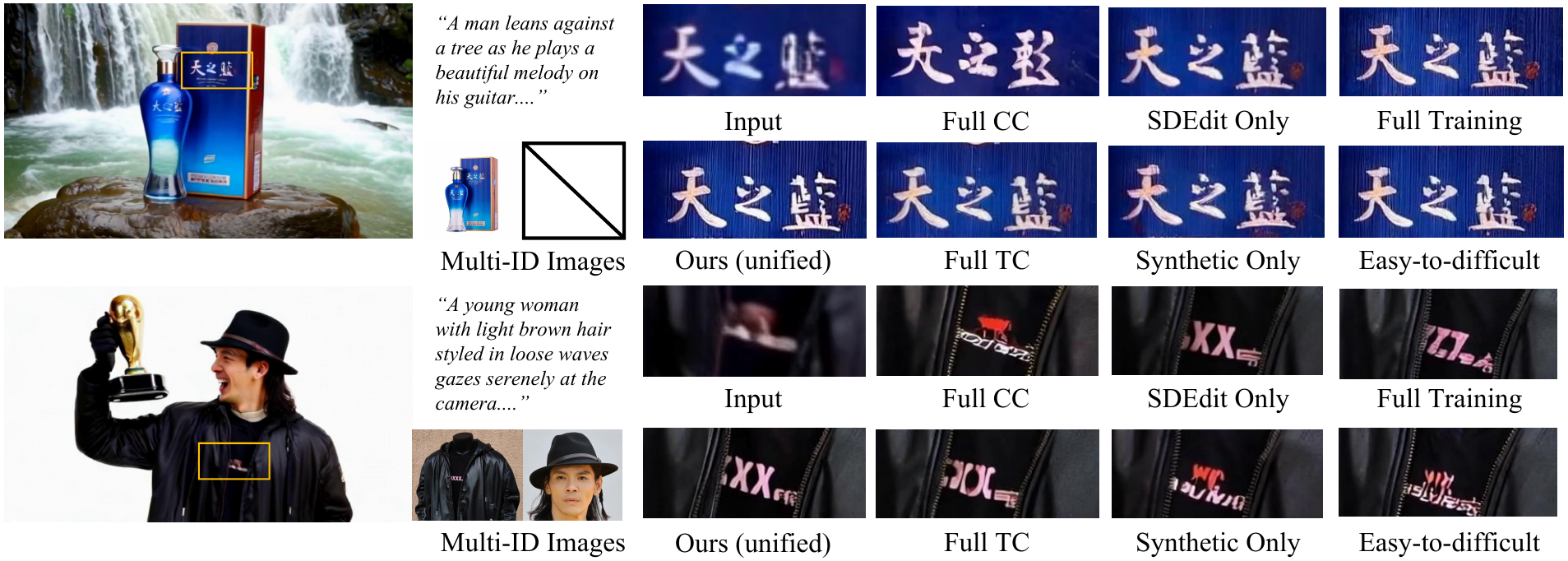}
    \caption{Qualitative comparisons with different components.}
    \label{fig:supp-ablation-study}
\end{figure}

Qualitative comparisons with different components are shown in Fig.~\ref{fig:supp-ablation-study}. For architecture design, full channel-concat (Full CC) struggles to inject visual references. For full token-concat (Full TC), although it achieves comparable results, it largely sacrifices the inference efficiency. For degradation effect, sdedit-only and synthetic-only methods lack in generating vivid details and preserving input ID images respectively. For training order, both full training and easy-to-difficult paradigm show suboptimal results compared with our difficult-to-easy paradigm, which demonstrates the effectiveness of our training strategy.

\subsubsection{4K Results}
\label{sec:supp-4k-results}
\begin{figure}[t]
    \centering
    \includegraphics[width=1.0\linewidth]{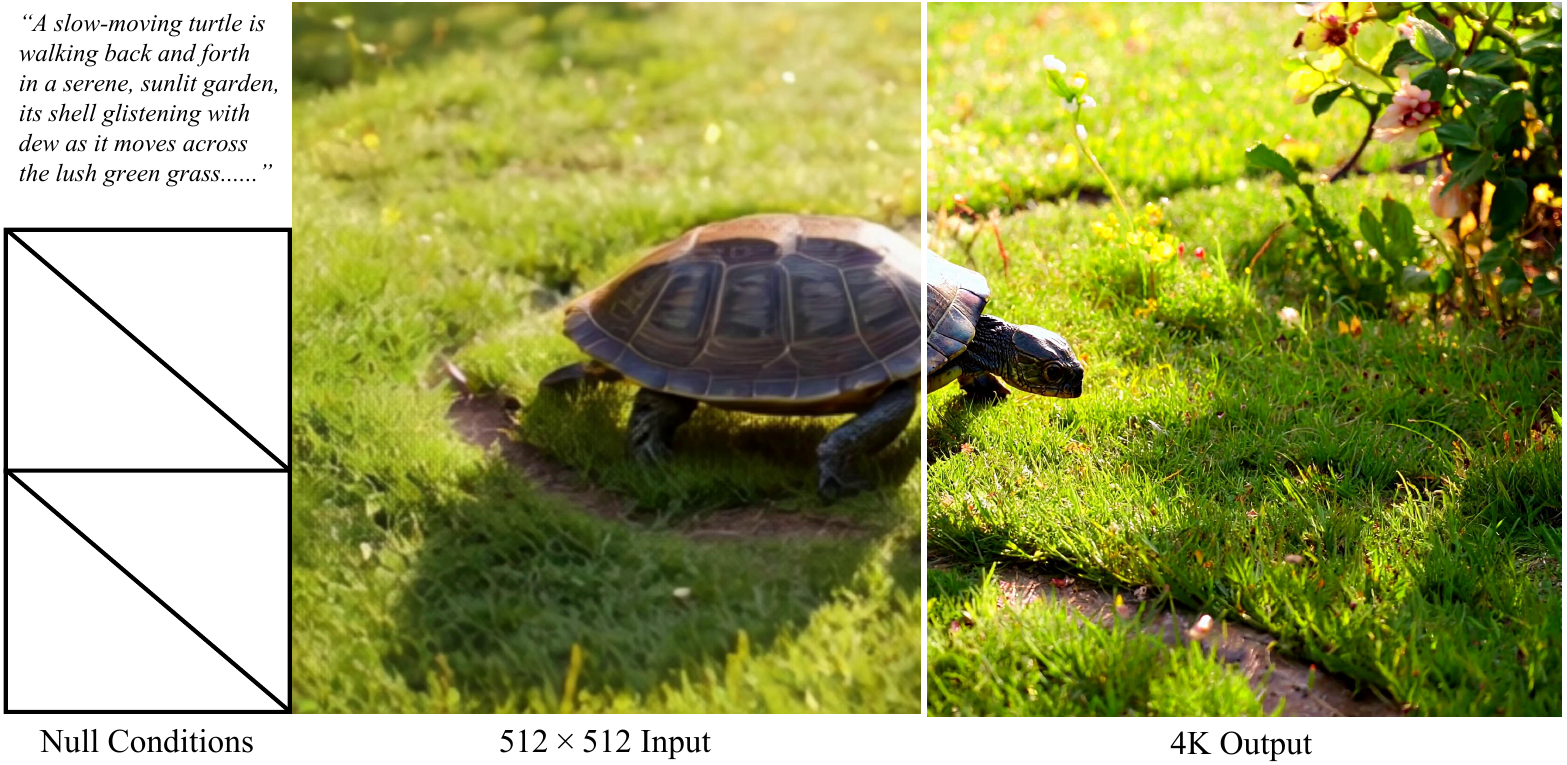}
    \caption{Additional 4K results on text-to-video generation task. \textbf{Zoom-in for best view.}}
    \label{fig:supp-4k-original}
\end{figure}

\begin{figure}[t]
    \centering
    \includegraphics[width=1.0\linewidth]{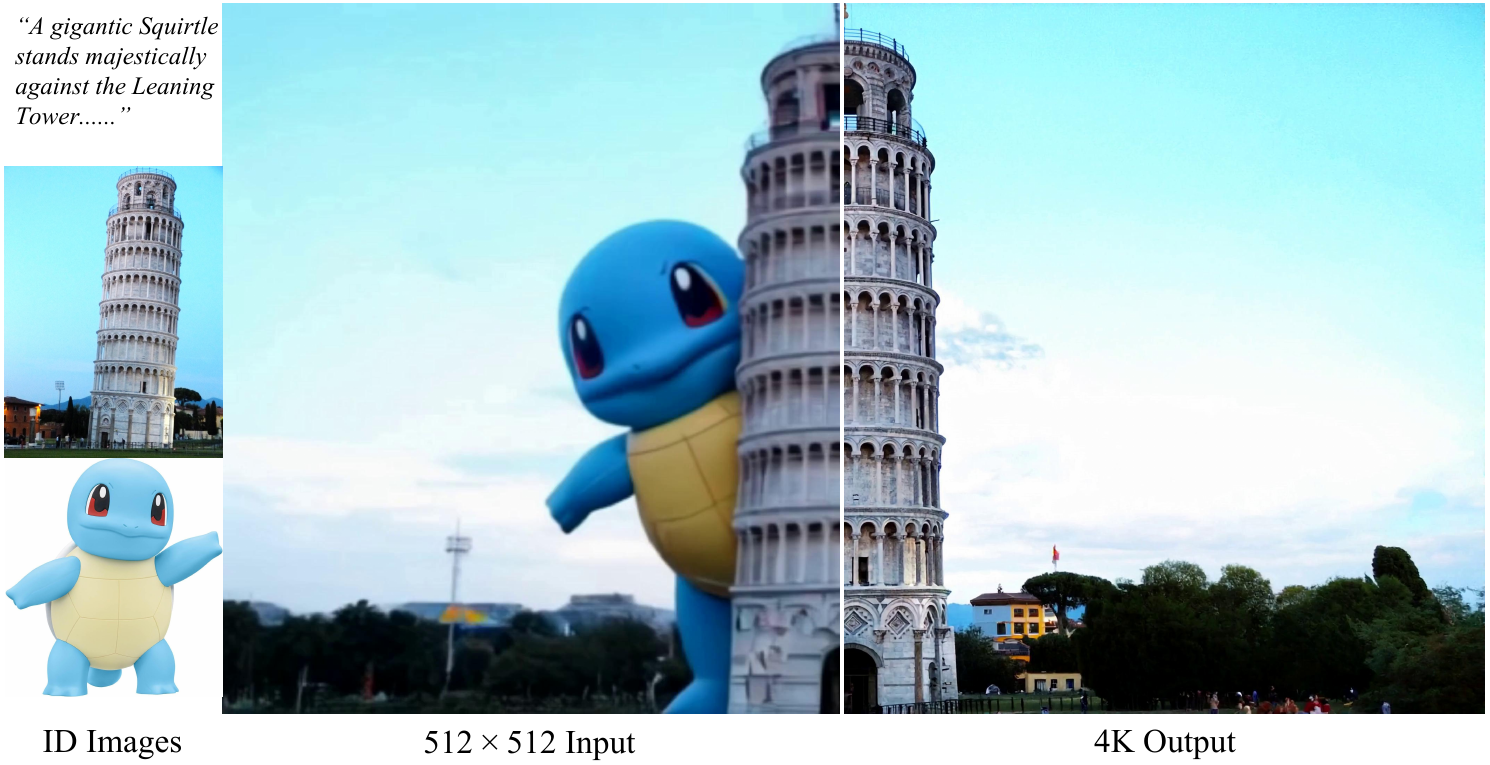}
    \caption{Additional 4K results on multi-ID image-guided text-to-video generation task. \textbf{Zoom-in for best view.}}
    \label{fig:supp-4k-id}
\end{figure}

\begin{figure}[t]
    \centering
    \includegraphics[width=1.0\linewidth]{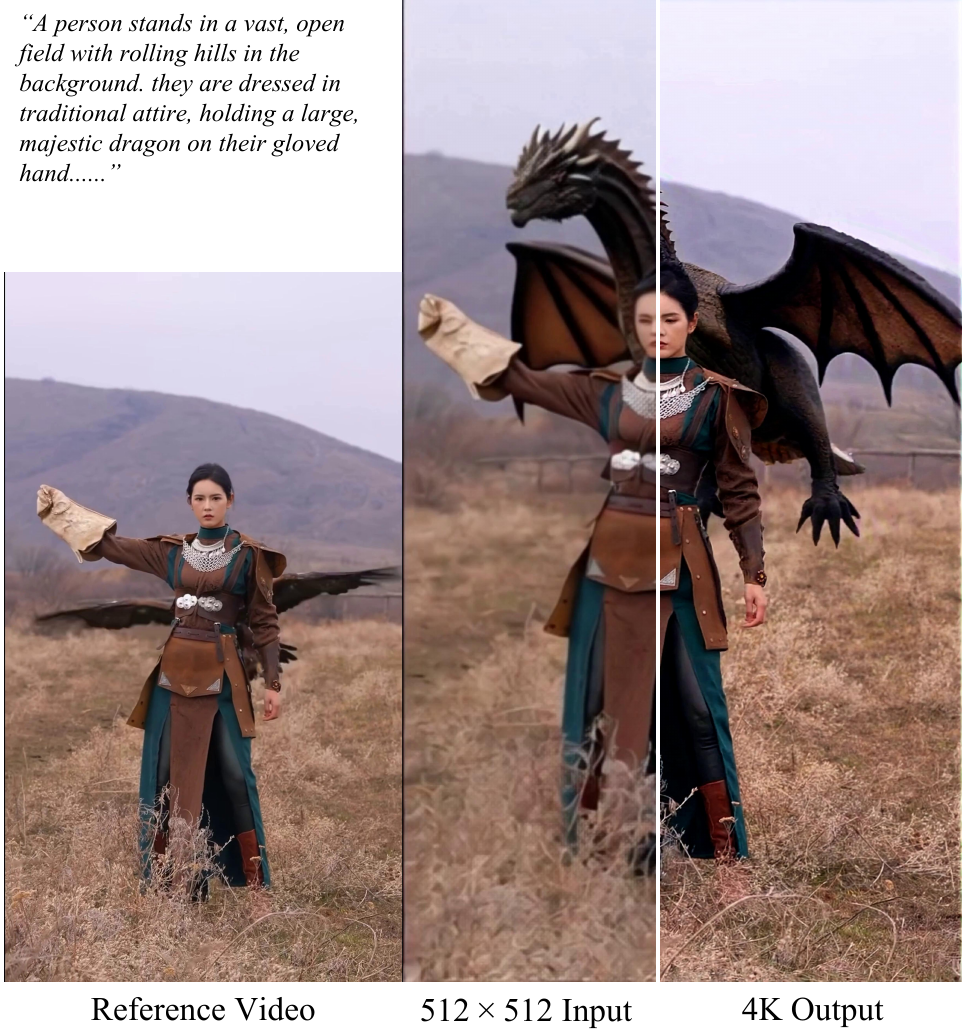}
    \caption{Additional 4K results on text-guided video editing task. \textbf{Zoom-in for best view.}}
    \label{fig:supp-4k-editing}
\end{figure}

Additional 4K results of text-to-video generation, multi-ID image-guided text-to-video generation and text-guided video editing tasks are presented in Fig.~\ref{fig:supp-4k-original}, ~\ref{fig:supp-4k-id} and ~\ref{fig:supp-4k-editing} respectively. The results show that the proposed cascaded framework excels at scaling resolution on all three controllable video generation tasks. We also present 4K videos in the supplementary material.

\end{document}

%% file: math_commands.tex
\usepackage{amsmath,amsfonts,bm}

\def\eqref#1{equation~\ref{#1}}
\def\1{\bm{1}}

\DeclareMathAlphabet{\mathsfit}{\encodingdefault}{\sfdefault}{m}{sl}
\SetMathAlphabet{\mathsfit}{bold}{\encodingdefault}{\sfdefault}{bx}{n}